\documentclass[lettersize,journal]{IEEEtran}
\usepackage{tabularx}
\usepackage{subcaption}
\usepackage{cite}
\usepackage{amsmath,amssymb,amsfonts}
\usepackage{algorithmic}
\usepackage{array}
\usepackage{graphicx}
\usepackage{textcomp}
\usepackage{xcolor}
\usepackage{booktabs}
\usepackage{textgreek}
\usepackage{tipa}
\usepackage{fancyhdr}
\usepackage{multirow}
\usepackage{color}
\usepackage{bbding}
\usepackage{stfloats}
\usepackage{algorithm}
\usepackage{hyperref}
\usepackage{cleveref}
\usepackage{url}
\usepackage{verbatim}
\usepackage{xpatch}

\usepackage{ifpdf}
 \ifpdf
 \else
 \fi

\def\BibTeX{{\rm B\kern-.05em{\sc i\kern-.025em b}\kern-.08em
    T\kern-.1667em\lower.7ex\hbox{E}\kern-.125emX}}
\usepackage{balance}

\begin{document}

\title{The Staged Knowledge Distillation in Video Classification: Harmonizing Student Progress by a Complementary Weakly Supervised Framework}
\author{Chao Wang, \emph{Member, IEEE} and Zheng Tang, \emph{Member, IEEE}
\thanks{Manuscript created March, 2023; Chao Wang is with China Academy of Railway Sciences, Beijing 100081, China (e-mail: chaowangasaph@outlook.com). Zheng Tang is with NVIDIA, Redmond, WA 98052, USA (e-mail: tangzhengthomas@gmail.com). Accepted as a Transactions Paper of IEEE Transactions on Circuits and Systems for Video Technology (TCSVT) in 2023.}}

\markboth{Journal of \LaTeX\ Class Files,~Vol.~18, No.~9, September~2020}%
{The Staged Knowledge Distillation in Video Classification: Harmonizing Student Progress by a Complementary Weakly Supervised Framework}

\maketitle

%%%%%%%%% ABSTRACT
\begin{abstract}
In the context of label-efficient learning on video data, the distillation method and the structural design of the teacher-student architecture have a significant impact on knowledge distillation. However, the relationship between these factors has been overlooked in previous research. To address this gap, we propose a new weakly supervised learning framework for knowledge distillation in video classification that is designed to improve the efficiency and accuracy of the student model. Our approach leverages the concept of substage-based learning to distill knowledge based on the combination of student substages and the correlation of corresponding substages. We also employ the progressive cascade training method to address the accuracy loss caused by the large capacity gap between the teacher and the student. Additionally, we propose a pseudo-label optimization strategy to improve the initial data label. To optimize the loss functions of different distillation substages during the training process, we introduce a new loss method based on feature distribution. We conduct extensive experiments on both real and simulated data sets, demonstrating that our proposed approach outperforms existing distillation methods in terms of knowledge distillation for video classification tasks. Our proposed substage-based distillation approach has the potential to inform future research on label-efficient learning for video data.
% The type of distillation method and the teacher-student architecture profoundly impact knowledge distillation in video classification. However, the internal relationship between them has been largely overlooked. Taking a novel approach to simulating human learning, we propose a more practical, weakly supervised learning framework for knowledge distillation and uncover the relationship between the type of distillation method and the structural design of the teacher-student architecture. Specifically, we introduce the concept of substage-based learning, which distills knowledge based on the combination of student substages and the correlation of corresponding substages. To enhance the student's training accuracy, we employ the progressive cascade training method to prevent the accuracy loss caused by the large gap between the teacher and the student while designing a pseudo-label optimization strategy to improve the initial data label. We optimize the loss functions of different distillation substages during the training process to fit our framework. We conduct extensive experiments on the CIFAR-100 and ImageNet datasets with various configurations, demonstrating that our proposed approach outperforms existing distillation methods in terms of knowledge distillation for video classification tasks. Our substage-based distillation approach has great potential. We hope it will benefit future research on the teacher-student relationship of knowledge distillation based on a weakly supervised video classification model.
\end{abstract}


%%%%%%%%% BODY TEXT
\section{Introduction}
\label{sec:intro}

In the past decade, deep learning, especially convolutional neural networks (CNNs), has achieved tremendous success in computer vision. In contrast, the weak supervision paradigm \cite{IEEEexample:2022Solve,IEEEexample:Liu2020DenserNetWS,IEEEexample:2020DenserNet,IEEEexample:Zhang2022ExploitingCA} has permeated every corner of its latest advances and has been widely used in the industrial sector. Image classification \cite{IEEEexample:classfy02,IEEEexample:Train,IEEEexample:Ke2022TowardsR3,IEEEexample:mao2022towards} is a fundamental task of computer vision, which gives rise to many other video tasks, such as video segmentation \cite{IEEEexample:Wang2021ASO,IEEEexample:Wu2022MultiLevelRL,IEEEexample:Liang2022LocalGlobalCA}, video instance segmentation \cite{IEEEexample:Liu2021SGNetSG,IEEEexample:Qin2023CoarsetoFineVI,IEEEexample:Wang2022LearningES}, object tracking \cite{IEEEexample:Dong2022AdaptiveST,IEEEexample:Shen2019DistilledSN,IEEEexample:Qin2023MotionTrackLR}, and visual recognition\cite{IEEEexample:Wang2022VisualRW,IEEEexample:Liang2022ATV,IEEEexample:Yan2022VideoCU,IEEEexample:Yan2022GLRGGR}. To achieve state-of-the-art performance, various CNN models have become increasingly deeper and wider, which require more computational power and memory for model inference. However, large-scale models are not ideal for industrial applications, making them challenging to be deployed in video processing applications \cite{IEEEexample:Tang2018SingleCameraAI,IEEEexample:Tang2019CityFlowAC,IEEEexample:Tang2019PAMTRIPM,IEEEexample:Bigdata,IEEEexample:Guo2022AVR,IEEEexample:Liu2020VisualLF,IEEEexample:Wang2022DeepPM,IEEEexample:Cheng2023FusionIN} such as self-driving cars and embedded systems. To tackle this issue, knowledge distillation (KD) has emerged as a promising technique to transfer knowledge from a larger and more complex teacher model to a smaller and less complex student model \cite{IEEEexample:knowledgedistillation01,IEEEexample:knowledgedistillation02,IEEEexample:knowledgedistillation04,IEEEexample:knowledgedistillation05}.

\begin{figure}[t]
  \begin{subfigure}{0.95\linewidth}
    \centering
    %\fbox{\rule{0pt}{2in} \rule{.9\linewidth}{0pt}}
    \includegraphics[width=1\linewidth]{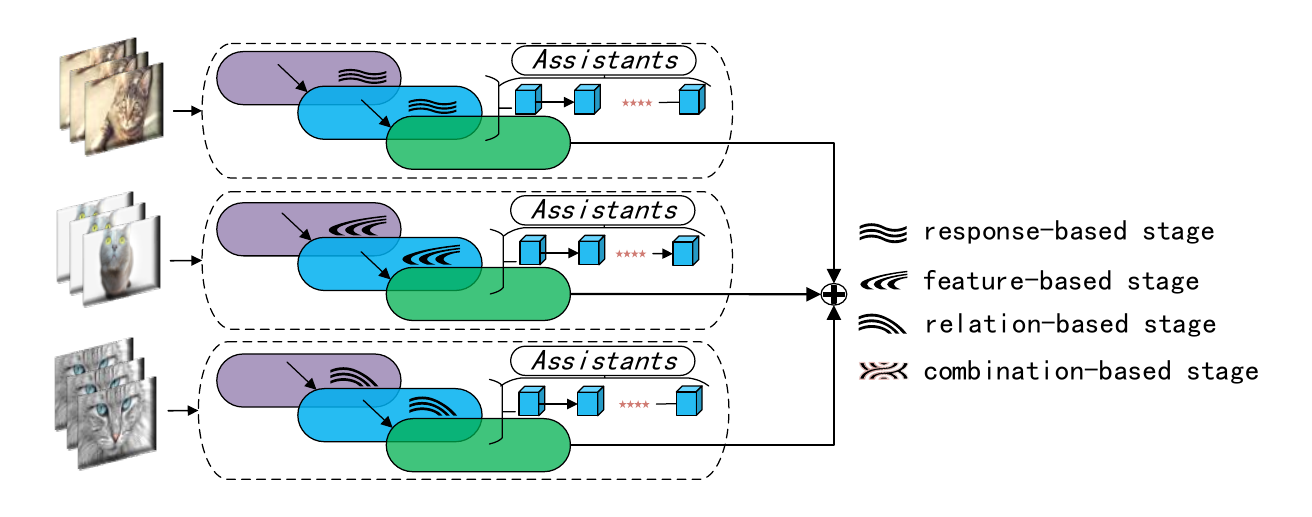}
    \caption{Staged Knowledge Distillation (SKD).}
    \label{fig:one-a}
  \end{subfigure}
  \vfill
  \begin{subfigure}{0.95\linewidth}
    %\fbox{\rule{0pt}{2in} \rule{.9\linewidth}{0pt}}
    \centering
    \includegraphics[width=1\linewidth]{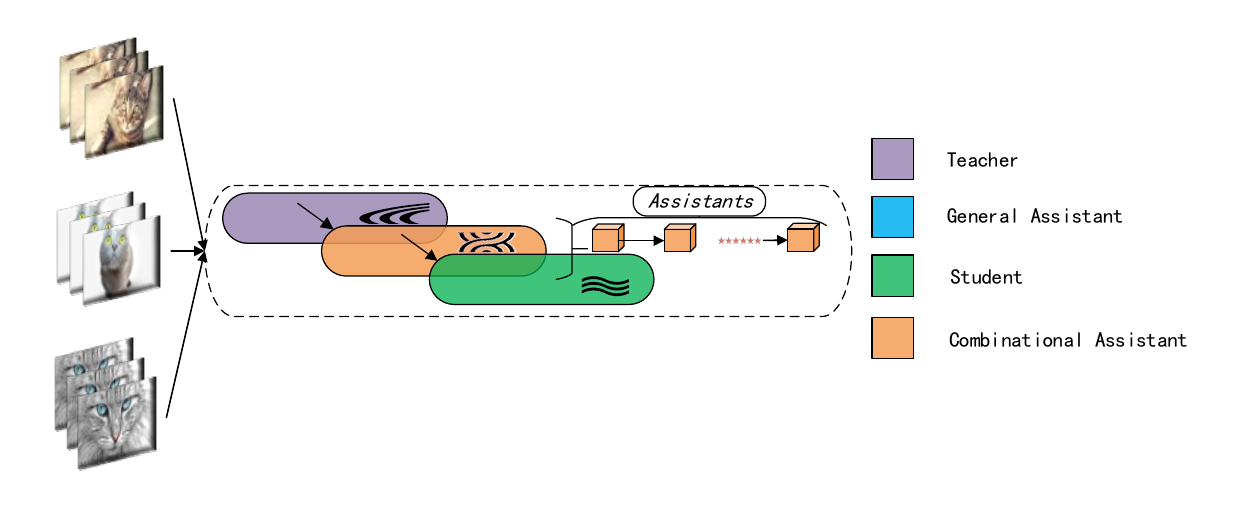}
    \caption{Relevant Staged Knowledge Distillation (RSKD).}
    \label{fig:one-b}
  \end{subfigure}
  \label{fig:one}
  \caption{Architecture diagram of our proposed method. (a) SKD strategy. (b) RSKD strategy. The purple, blue, orange and green colors denote the teacher, general teaching assistant, combined teaching assistant and student, respectively. The shape of the line denotes the type of substage.}
  \label{fig:one}
\end{figure}

The field of KD has seen significant progress in recent years, with research focusing on the types of KD, loss functions, and the design of teacher-student architectures \cite{IEEEexample:survey}. Different types of KD, including response-based \cite{IEEEexample:survey}, feature-based \cite{IEEEexample:knowledgedistillation02,IEEEexample:knowledgedistillation04,IEEEexample:FT}, and relation-based \cite{IEEEexample:knowledgedistillation05,IEEEexample:AN}, have been proposed to transfer knowledge from the teacher to the student. Researchers have also proposed various loss functions to enhance the student's performance, such as distilling knowledge with dense representations \cite{IEEEexample:DKD}, attention-based knowledge distillation \cite{IEEEexample:AB}, and temperature-damping normalization \cite{IEEEexample:TDN}. Furthermore, different teacher-student architectures have been designed, including multiple teachers-students \cite{IEEEexample:FEED,IEEEexample:HES,IEEEexample:LFM,IEEEexample:BA,IEEEexample:MC,IEEEexample:DPR,IEEEexample:ASS}, self-training \cite{IEEEexample:RKD,IEEEexample:ST}, and mutual learning \cite{IEEEexample:DML}. However, despite the abundance of research on the topic, there has been little attention paid to the internal relationship between the type of distillation method and the structural design of the teacher-student architecture.
% In the field of KD, three research directions\cite{IEEEexample:survey} play a pivotal role: types of KD, loss functions, and the design of teacher-student architectures. The types of KD mainly include response-based\cite{IEEEexample:survey}, feature-based\cite{IEEEexample:knowledgedistillation02,IEEEexample:knowledgedistillation04,IEEEexample:FT}, and relation-based\cite{IEEEexample:knowledgedistillation05,IEEEexample:AN}. The loss enhances the student model's performance by reformulating the KD loss function\cite{IEEEexample:DKD,IEEEexample:AB,IEEEexample:ACO,IEEEexample:TDN}. The teacher-student architecture is divided into multiple teachers-students\cite{IEEEexample:FEED,IEEEexample:HES,IEEEexample:LFM,IEEEexample:BA,IEEEexample:MC,IEEEexample:DPR,IEEEexample:ASS}, self-training\cite{IEEEexample:RKD,IEEEexample:ST}, and mutual learning\cite{IEEEexample:DML}. While most KD methods leverage various kinds of knowledge, they often overlook the combination of knowledge between teacher-student architectures.

KD is a process that mimics human learning by systematically transferring knowledge from a teacher model to a student model. To maximize the effectiveness of this process, we propose a novel framework called Staged Knowledge Distillation (\textbf{SKD}), which consists of three distinct learning stages, as shown in Figure \ref{fig:one-a}. We argue that a blind, single-stage approach may be inefficient and ineffective due to its increased cost and potential for bias.
% Intuitively, the KD process is analogous to human learning; mastering any knowledge necessitates a systematic learning stage initially. We hypothesize that combining learning stages facilitates students' comprehension of the teacher's instruction. Therefore, we have divided KD into three distinct stages. At the same time, we argue that the analogy with the human learning process implies that a single, blind learning stage may be inefficient and ineffective due to its increased cost and potential for bias. So, we propose a novel framework for KD: Staged Knowledge Distillation (\textbf{SKD}), a step-by-step approach from shallow to deep, using a multi-branch cascade structure, as illustrated in Figure \ref{fig:one-a}.

\begin{figure}
  \begin{subfigure}[t]{0.49\linewidth}
    \centering
    %\fbox{\rule{0pt}{2in} \rule{.9\linewidth}{0pt}}
    \includegraphics[width=1.05\textwidth,height=0.16\textheight]{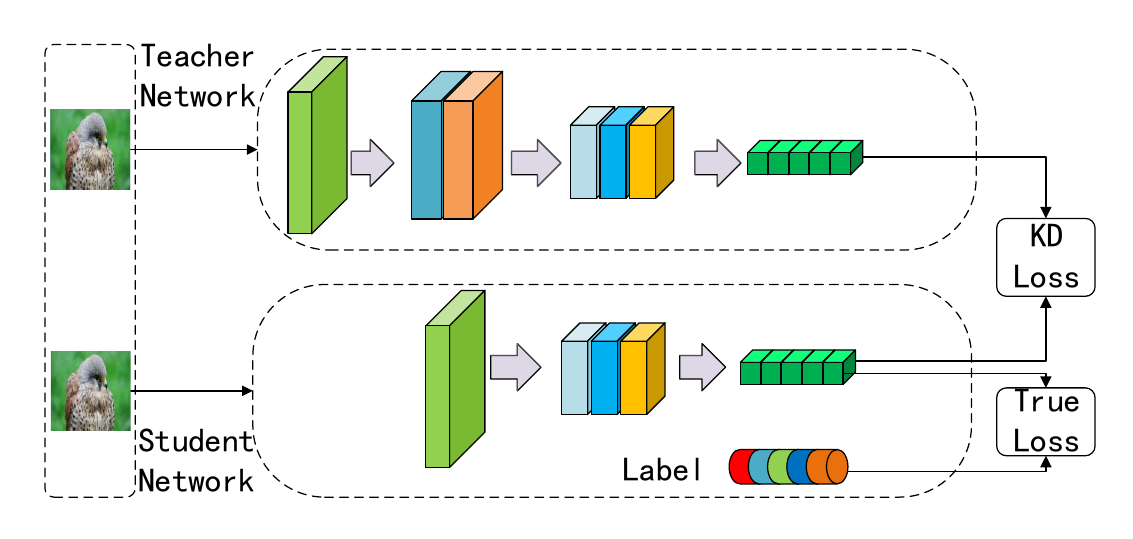}
    \caption{Response-Based Substage.}
    \label{fig:short-a}
  \end{subfigure}
  \hfill
  \begin{subfigure}[t]{0.49\linewidth}
    %\fbox{\rule{0pt}{2in} \rule{.9\linewidth}{0pt}}
    \centering
    \includegraphics[width=1.05\textwidth,height=0.16\textheight]{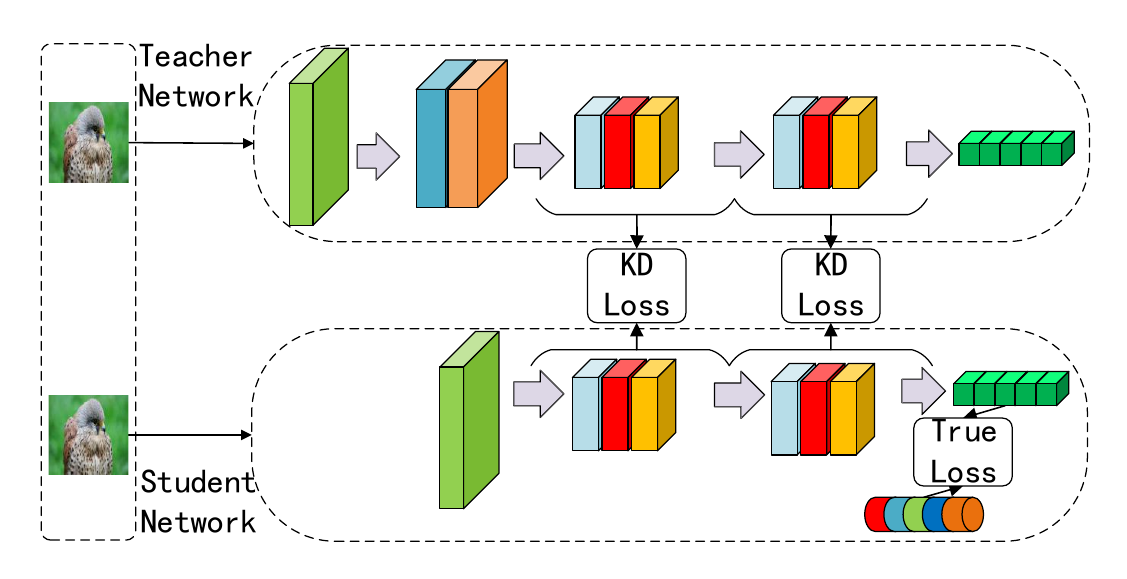}
    \caption{Relation-Based Substage.}
    \label{fig:short-b}
  \end{subfigure}
  \vfill
  \begin{subfigure}[t]{0.8\linewidth}
    %\fbox{\rule{0pt}{2in} \rule{.9\linewidth}{0pt}}
    \centering
    \includegraphics[width=1.3\textwidth]{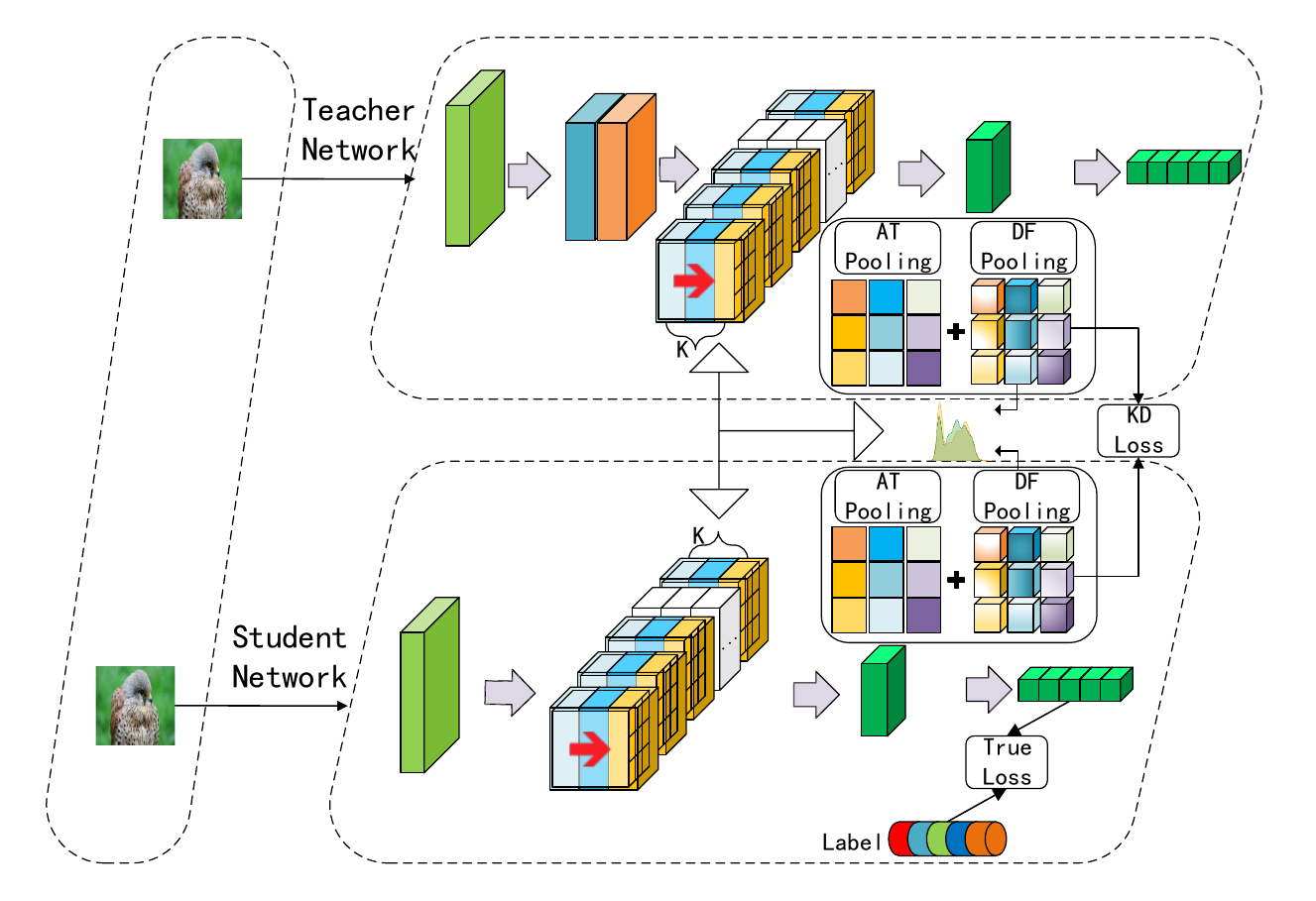}
    \caption{Feature-Based Substage.}
    \label{fig:short-c}
  \end{subfigure}
  \caption{Illustration of three types of substages. (a) Response-Based Substage directly mimics the final result of the teacher model. (b) Relation-Based Substage explores the relationship between feature mapping pairs. (c) Feature-Based Substage focuses on the output distillation process of the intermediate feature layers.}

  \label{fig:two}

\end{figure}

To address this issue, we adopt a parallel multi-branch structure training method divided into three substages (see Figure \ref{fig:two}), which emulates the human learning process and improves the algorithm's generalization ability. However, if the capacity gap between the teacher and the student is too large, the training effect will be diminished \cite{IEEEexample:ASS}. To overcome this limitation, we introduce the teaching assistant method \cite{IEEEexample:ASS}, which we refer to as cascade training. Additionally, we use a weakly supervised label enhancement technique that generates high-quality pseudo labels to supervise the classification in the video.
% First, we divide the distillation process into three substages in \cref{fig:two}, which emulate the human learning process and adopt a parallel multi-branch structure training method to maximize the effectiveness of each stage. However, if the capacity gap between the teacher and the student is too large, the training effect will be diminished \cite{IEEEexample:ASS}. To address this issue, we introduce the teaching assistant method in \cite{IEEEexample:ASS}, which we refer to as cascade training. To improve the algorithm's generalization ability, we also adopt a weakly supervised label enhancement technique, which uses the complementary representation of spatiotemporal signals to generate high-quality pseudo labels to supervise the classification in the video.

The final student model in SKD is obtained by combining the results of each stage, which reduces the model's generalization error. However, the same learning framework is used internally for each parallel training branch, leading to the excessive correlation between models \cite{IEEEexample:MOD} and increased inference time and memory occupation when deployed. To address this issue, we propose a novel variant called Relevant Staged Knowledge Distillation (\textbf{RSKD}), which employs a combination stage (\textbf{CS}) approach, as illustrated in Figure \ref{fig:one-b}. RSKD omits the ensemble process and instead uses a combination-based stage distillation.
% Second, for SKD, the final student is obtained by combining the results of each stage, which can reduce the model's generalization error. Ensemble models typically yield better results with more types of models \cite{IEEEexample:MOD}. However, the same learning framework is used internally for each parallel training branch, leading to excessive correlation between models and increased inference time and memory occupation when deployed. To address this issue, we propose a novel variant, Relevant Staged Knowledge Distillation (\textbf{RSKD}), which employs a combination stage (\textbf{CS}) approach to reduce the correlation between learning processes, as illustrated in Figure \ref{fig:one-b}. RSKD omits an ensemble process and instead adopts a combination-based stage distillation compared to SKD.

Finally, we enhance the KD loss function between different processes to adapt to our proposed SKD and RSKD frameworks. For feature-based KD, we propose an improved method for comparing the distribution of feature dimensions (\textbf{DF}), which selects the teacher's top K channels and the student's K channels for distillation based on the original loss function. This improvement can significantly improve the effectiveness of distillation.
% Finally, we enhance the KD loss function between different processes to adapt to our proposed SKD and RSKD frameworks. For feature-based KD, the attention mechanism in literature \cite{IEEEexample:knowledgedistillation04} converts the knowledge between the teacher and the student through compression in the channel dimension. We propose an improved method for comparing the distribution of feature dimensions (\textbf{DF}), which selects the teacher's top K channels and the student's K channels for distillation based on the original loss function. This improvement can significantly improve the effectiveness of distillation.

In this paper, we make several contributions to the field of knowledge distillation:
\begin{itemize}
    \item We propose a new loss method, DF, which is based on the feature distribution and is able to uncover knowledge hidden in the distribution of features.
    \item We address traditional challenges in knowledge distillation by treating it as a stage learning problem and proposing a combination stage approach (CS).
    \item We propose two new weakly supervised distillation frameworks. The first, SKD, is a substage-integrated framework that simulates the human-stage learning process. The second, RSKD, is based on a relevant combination stage approach that improves efficiency and accuracy.
    \item We conducted extensive experiments on two analog data sets (CIFAR-100 and ImageNet) and one real data set (UCF101) \cite{IEEEexample:Soomro2012UCF101AD}, and demonstrate that our models achieve state-of-the-art and competitive results in video classification using knowledge distillation.
\end{itemize}
These contributions demonstrate the effectiveness of our proposed methods and provide insights into the potential of knowledge distillation for improving video classification models.
% Overall, our main contributions in this paper are as follows:
% \begin{itemize}
%     \item A new loss method based on feature distribution, DF, is proposed. This method can uncover the knowledge hidden in the feature distribution.
%     \item We address challenging traditional problems by treating KD as a stage learning problem and propose a combination stage approach, called CS, as a resolution.
%     \item Two new weakly supervised distillation frameworks are proposed. First, we propose a substage integrated framework SKD to simulate the human-stage learning process. Second, we propose RSKD based on the relevant combination stage to improve efficiency and accuracy.
%     \item We conduct extensive experiments of our framework on two benchmark datasets: CIFAR-100 and ImageNet, demonstrating that our models achieve state-of-the-art or competitive results in KD video classification.
% \end{itemize}

%-------------------------------------------------------------------------
\section{Related Works}
\label{sec:formatting}

\begin{figure*}
  \begin{subfigure}{0.48\linewidth}
    \centering
    %\fbox{\rule{0pt}{2in} \rule{.9\linewidth}{0pt}}
    \includegraphics[width=1\textwidth]{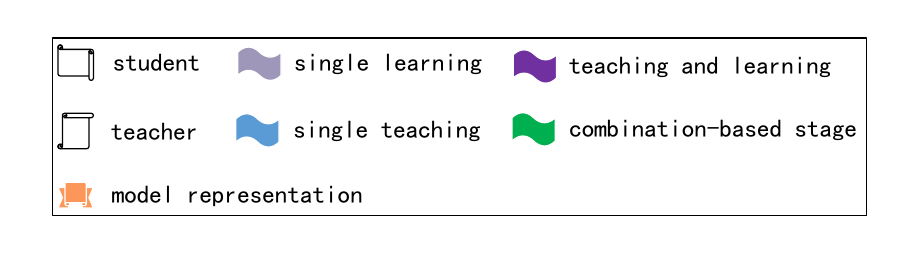}
    \label{fig:three-a}
  \vspace{-1.9cm}
  \end{subfigure}
  \vfill
  \begin{subfigure}{0.1\linewidth}
    \centering
    %\fbox{\rule{0pt}{2in} \rule{.9\linewidth}{0pt}}
    \includegraphics[width=0.7\textwidth]{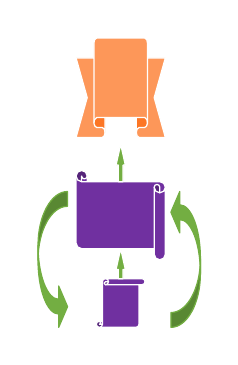}
    \caption{ST.\cite{IEEEexample:RKD}}
    \label{fig:three-a}
  \end{subfigure}
  \hfill
  \begin{subfigure}{0.15\linewidth}
    %\fbox{\rule{0pt}{2in} \rule{.9\linewidth}{0pt}}
    \centering
    \includegraphics[width=0.7\textwidth]{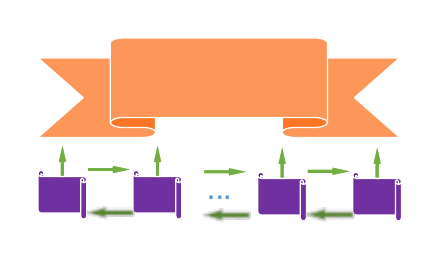}
    \caption{ML.\cite{IEEEexample:DML}}
    \label{fig:three-b}
  \end{subfigure}
  \hfill
  \begin{subfigure}{0.15\linewidth}
    %\fbox{\rule{0pt}{2in} \rule{.9\linewidth}{0pt}}
    \centering
    \includegraphics[width=0.7\textwidth]{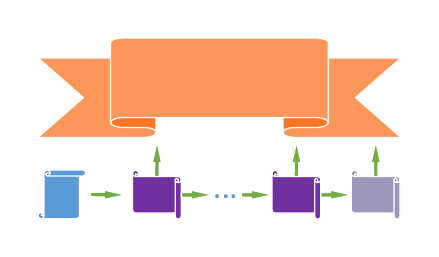}
    \caption{OTM.\cite{IEEEexample:BA}}
    \label{fig:three-c}
  \end{subfigure}
  \hfill
  \begin{subfigure}{0.22\linewidth}
    %\fbox{\rule{0pt}{2in} \rule{.9\linewidth}{0pt}}
    \centering
    \includegraphics[width=0.7\textwidth]{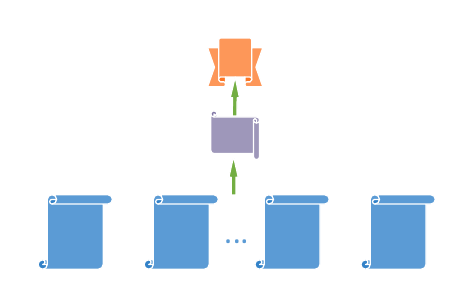}
    \caption{MTO.\cite{IEEEexample:MC}\cite{IEEEexample:LFM}\cite{IEEEexample:HES}\cite{IEEEexample:FEED}}
    \label{fig:three-d}
  \end{subfigure}
  \hfill
  \begin{subfigure}{0.09\linewidth}
    %\fbox{\rule{0pt}{2in} \rule{.9\linewidth}{0pt}}
    \centering
    \includegraphics[width=0.7\textwidth]{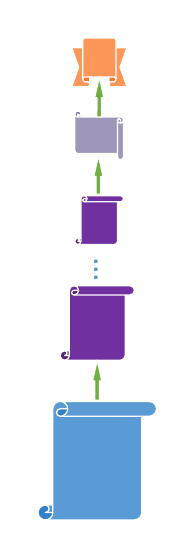}
    \caption{AT.\cite{IEEEexample:ASS}}
    \label{fig:three-e}
  \end{subfigure}
  %\hfill
  %\begin{subfigure}{0.15\linewidth}
   % \centering
    %\includegraphics[width=0.9\textwidth]{3f.pdf}
   % \caption{SKD.}
    %\label{fig:three-f}
  %\end{subfigure}
  %\hfill
  %\begin{subfigure}{0.08\linewidth}
    %\centering
    %\includegraphics[width=0.9\textwidth]{3g.pdf}
    %\caption{RSKD.}
    %\label{fig:three-g}
  %\end{subfigure}
  \caption{Five categories of teacher-student network design. The orientation represents model type: teacher or student; The color indicates function: teach or learn; The figure size represents model size; The final model is in orange.}
  \label{fig:three}
\vspace{-0.6cm}
\end{figure*}

\subsection{Weakly-Supervised Paradigms}

Automatic video detection has gained increasing attention due to its usefulness in various intelligent surveillance systems. In recent years, research has focused on the weakly supervised learning paradigms due to the scarcity of clip-level annotations in video datasets \cite{IEEEexample:2021MIST,IEEEexample:ShuoLi2023SelfTrainingML,IEEEexample:2021Weakly,IEEEexample:2019Temporal,IEEEexample:2019Motion}.
% Detecting videos automatically is gaining more attention due to its versatility in various intelligent surveillance systems. Scarce irregularities in videos have led recent investigations to chiefly concentrate on the weakly supervised learning paradigm \cite{IEEEexample:2021MIST,IEEEexample:ShuoLi2023SelfTrainingML,IEEEexample:2021Weakly,IEEEexample:2019Temporal,IEEEexample:2019Motion}.

Existing one-stage methods \cite{IEEEexample:2021Weakly,IEEEexample:2019Temporal,IEEEexample:2019Motion} rely on Multiple Instance Learning to describe anomaly detection as a regression problem and adopt an ordering loss approach. However, the lack of clip-level annotations often results in low accuracy. To address this issue, a two-stage approach based on self-training has been proposed \cite{IEEEexample:2021MIST,IEEEexample:ShuoLi2023SelfTrainingML}. This approach generates pseudo-labels for clips and uses them to refine the discriminative representation. However, the generated pseudo-labels' uncertainty is not considered, which may negatively impact the classifier's performance.
% Existing one-stage methods \cite{IEEEexample:2021Weakly,IEEEexample:2019Temporal,IEEEexample:2019Motion} that rely on Multiple Instance Learning describe anomaly detection as a regression problem and adopt the ordering loss approach. However, these methods are often inaccurate due to the lack of clip-level annotations. A two-stage approach based on self-training\cite{IEEEexample:2021MIST,IEEEexample:ShuoLi2023SelfTrainingML} has been proposed to address this issue. This approach generates pseudo-labels for clips and then uses them to refine the discriminant representation. However, the generated pseudo-labels' uncertainty is not considered, which may negatively impact the classifier's performance.

\subsection{Types of Knowledge Distillation}

Knowledge distillation (KD) was first proposed in \cite{IEEEexample:Modelcompression01} and has since been widely used to transfer knowledge from a large teacher network to a smaller student network \cite{IEEEexample:knowledgedistillation01}. Direct response-based distillation methods use softened label outputs to match the teacher and student predictions. However, they can be challenging to converge in some cases, leading to the development of feature-based distillation methods.
% Knowledge distillation was first proposed in \cite{IEEEexample:Modelcompression01} and has since been widely used in \cite{IEEEexample:knowledgedistillation01} to transfer dark knowledge from a large and computationally expensive teacher network to a single, smaller student network by using softened label outputs. However, this direct response-based distillation method can be challenging to converge in some cases, leading to the development of the feature-based distillation method.

FitNets \cite{IEEEexample:knowledgedistillation02} proposed a two-stage method based on the middle layer to improve the performance of student networks. However, introducing an intermediate layer for dimension conversion may introduce additional parameters. Attention Transfer \cite{IEEEexample:knowledgedistillation04} introduced the attention mechanism to transform the feature information of each layer inside the teacher network to the student network.
% FitNets \cite{IEEEexample:knowledgedistillation02} first use the two-stage method based on the middle layer, which can train deeper networks to improve the performance of student networks. However, due to the mismatch between the teacher and student network dimensions, it is necessary to introduce an intermediate layer for dimension conversion, which introduces additional parameters. Attention Transfer \cite{IEEEexample:knowledgedistillation04} introduces the attention mechanism into KD for the first time, which can transform the feature information of each layer inside the teacher network to the student network.

Relation-based distillation explores the relationship between different layers in the teacher and student networks. In \cite{IEEEexample:knowledgedistillation05}, the authors define the matrix correlation to measure the feature correlations between input and output layers of teachers and students. This method requires teachers and students to have the same structure, which limits the model's generalization ability.
% Since response-based and feature-based both use the output results of a specific layer of the teacher network for distillation and lack attention to the transferred knowledge from the global structure, relation-based is used to explore the relationship between different layers. The authors of \cite{IEEEexample:knowledgedistillation05} define the matrix correlation to measure the feature correlations between input and output layers of teachers and students so that they can be as similar as possible and then continue training for the original task. However, this method requires teachers and students to have the same structure, which limits the model's generalization ability.

\subsection{Structures of Knowledge Distillation}

KD's teacher-student network structure design can be summarized into five categories, as shown in \cref{fig:three}.

The self-training method \cite{IEEEexample:RKD} belongs to \cref{fig:three-a}, where a poor teacher is used to train the student network. The teacher is trained to bypass the soft labels of the student through the label smoothing technique.

The authors of \cite{IEEEexample:DML} propose a deep mutual learning strategy under \cref{fig:three-b}, in which a set of student networks learns and guides each other throughout the training process.

Instead of compressing the model, the student network \cite{IEEEexample:BA} is trained with the same parameterization as the teacher network. The final result is obtained by combining the training results of multiple students and calculating their averages, falling under ensemble learning as shown in \cref{fig:three-c}.

As demonstrated in \cite{IEEEexample:LFM, IEEEexample:MC}, a model with multiple teachers and a single student is proposed to reduce the complexity of the student network. Subsequently, a novel collaborative teaching strategy \cite{IEEEexample:HES} is proposed, leveraging the knowledge of two remarkable teachers to discover valuable information. Park \cite{IEEEexample:FEED} proposes a feature-based ensemble training method for KD, which employs multiple nonlinear transformations to transfer the knowledge of multiple teachers in parallel. These approaches can be visualized in \cref{fig:three-d}.

However, the primary issue with these approaches is that they rely on a single KD type, which can lead to a weakened distillation effect if there is a significant discrepancy between the teacher and student models. To address this, a novel distillation method for teaching assistants is proposed \cite{IEEEexample:ASS}, as shown in \cref{fig:three-e}, which addresses the decreased KD efficiency when the gap between the teacher and student networks is too large. However, the lack of multi-type KD in stages results in a weak generalization ability of the model, making it challenging to apply to large-scale datasets and models.

%-------------------------------------------------------------------------
\section{Proposed Method}

In this section, we present our proposed method for knowledge distillation, which consists of three key components. First, we improve the loss function of the distillation substage to optimize the distillation process and make it adaptable to our proposed method. Second, we analyze the effectiveness of multi-branch stage distillation and propose a novel approach that combines distillation substages with less correlation within the structure to achieve maximum distillation effect. Finally, we introduce two new distillation methods, SKD and RSKD, which aim to simulate the learning process of the human stage as closely as possible by applying the substage and multi-branch combined distillation method. Together, our proposed method provides a comprehensive framework for efficient and effective knowledge distillation.
% We will first improve the loss function of the KD substage to optimize the distillation process and make it adaptable to the method in this paper. We then analyze multi-branch stage distillation and combine the distillation substages with less correlation within the structure to achieve the maximum distillation effect. We finally propose two distillation methods, SKD and RSKD, to simulate the learning process of the human stage as much as possible by applying the substage and multi-branch combined distillation method.

\subsection{Label Generation}

To simulate a video classification dataset, we designed a composite-label generator inspired by the parallel multi-head classifier in \cite{IEEEexample:Zhang2022ExploitingCA}. The generator consists of an original-label generator and a pseudo-label generator. The main idea is to utilize a CycleGAN \cite{IEEEexample:Rai2018UnpairedIT} to generate labels with relatively high approximate probabilities and then use random selection to generate labels with relatively low approximate probabilities (i.e., anomalous labels), and perform interpolation combinations.
% Inspired by the design of a parallel multi-head classifier in \cite{IEEEexample:Zhang2022ExploitingCA} to compose a label generator, we design a composite-label generator consisting of an original-label generator and a pseudo-label generator to construct a video classification data set. 

\begin{figure*}
  \begin{subfigure}{0.314\linewidth}
    \centering
    %\fbox{\rule{0pt}{2in} \rule{.9\linewidth}{0pt}}
    \includegraphics[width=1\textwidth]{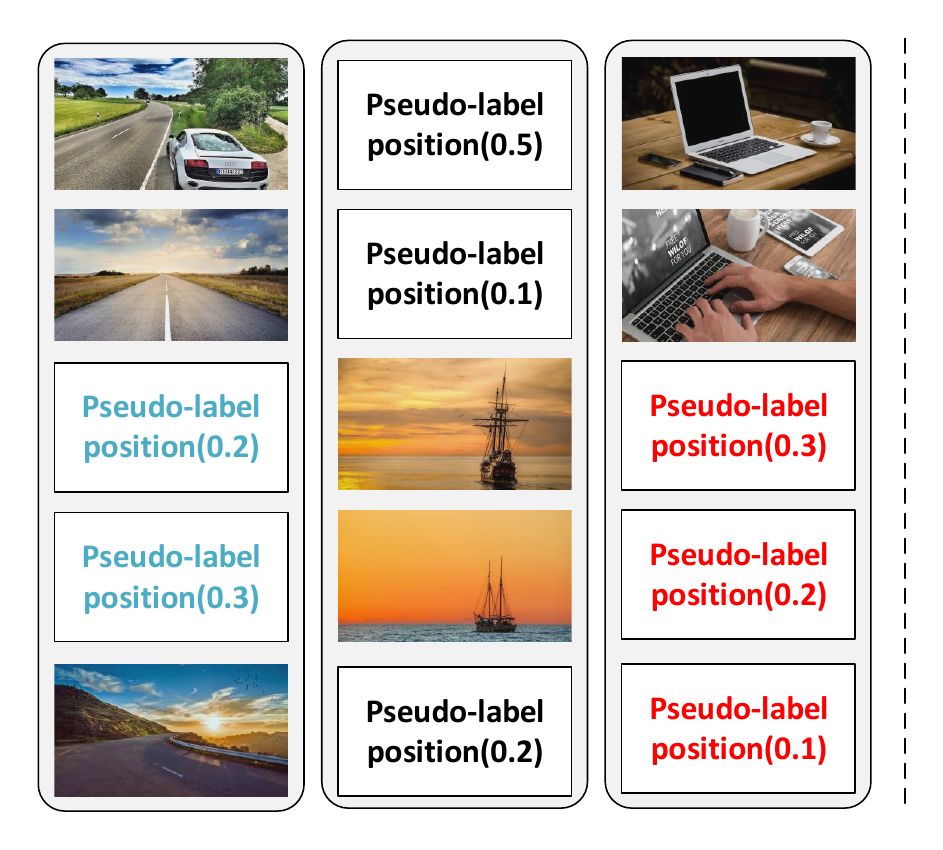}
    \caption{Original-label image.}
    \label{fig:six-a}
  \end{subfigure}
  \hfill
  \begin{subfigure}{0.361\linewidth}
    %\fbox{\rule{0pt}{2in} \rule{.9\linewidth}{0pt}}
    \centering
    \includegraphics[width=1\textwidth]{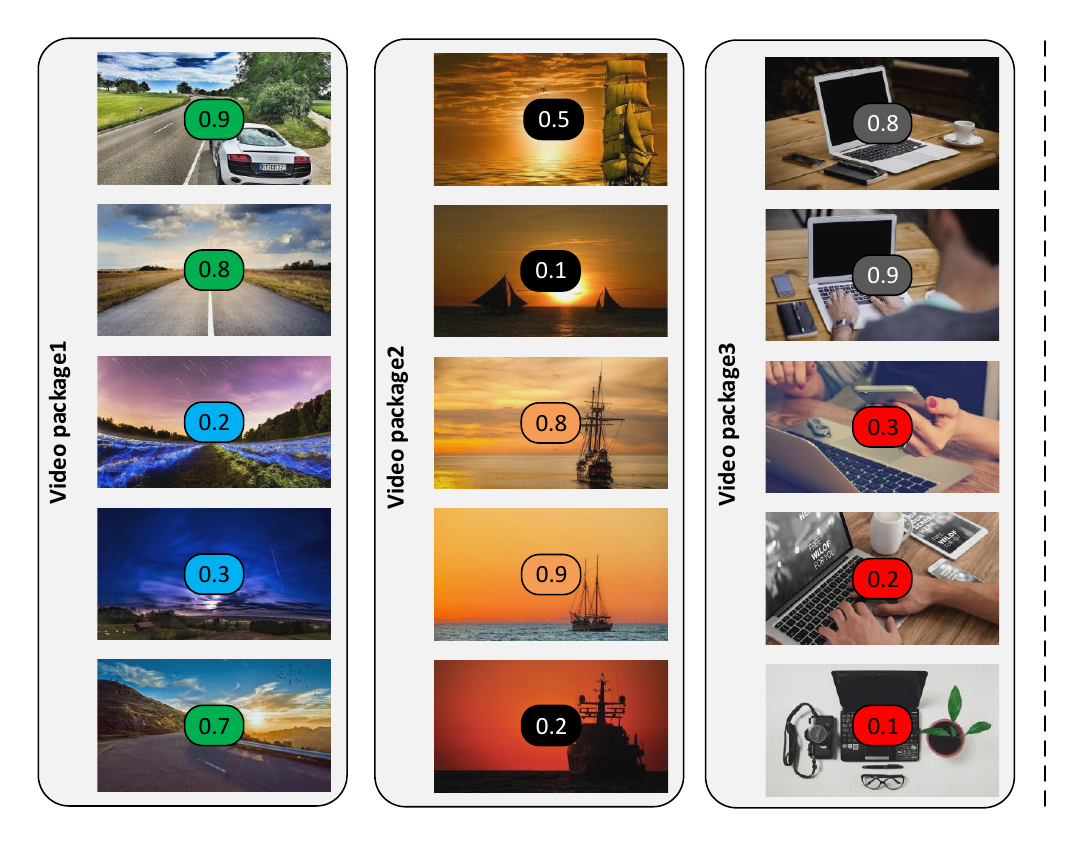}
    \caption{Composite-label image.}
    \label{fig:six-b}
  \end{subfigure}
  \hfill
  \begin{subfigure}{0.305\linewidth}
    %\fbox{\rule{0pt}{2in} \rule{.9\linewidth}{0pt}}
    \centering
    \includegraphics[width=1\textwidth]{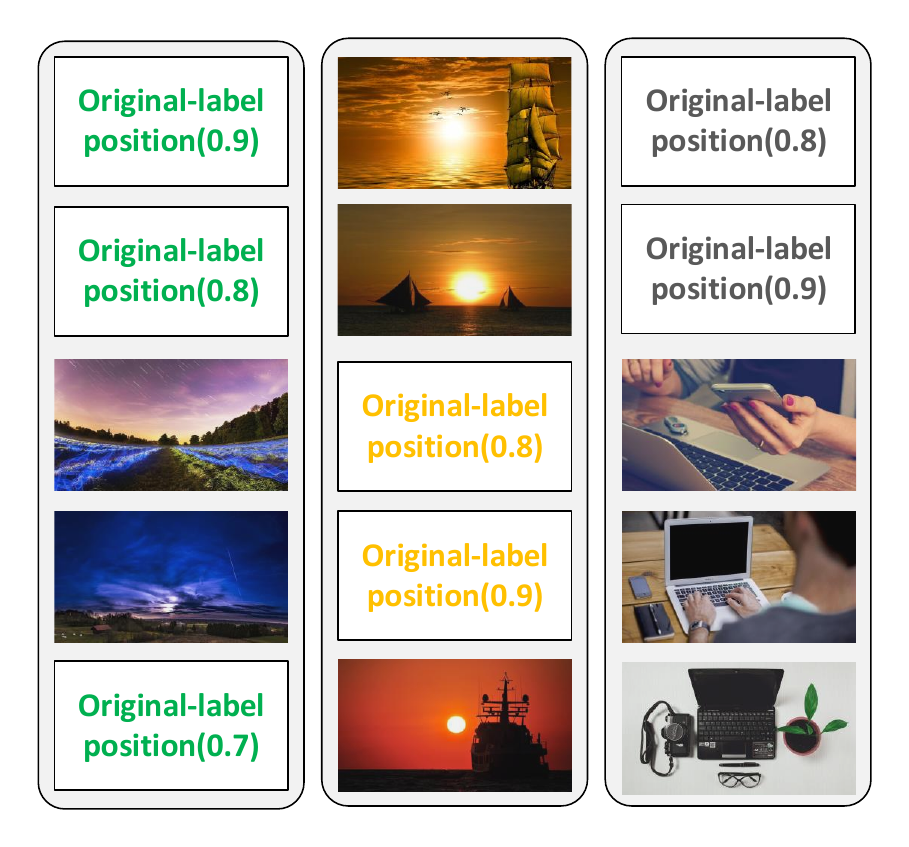}
    \caption{Pseudo-label image.}
    \label{fig:six-c}
  \end{subfigure}
  \caption{The proposed process for generating pseudo-labels. The input image is fed into a pre-training model to obtain the original image label and its uncertainty. A reliable clip is selected based on this uncertainty, and pseudo-labels are generated for this clip using random search and the CycleGAN\cite{IEEEexample:Rai2018UnpairedIT} method. The uncertainty of the pseudo-labels is calculated, and clip-level pseudo-labels are selected based on this uncertainty. The reliable and pseudo-labels are then combined to create a comprehensive video classification label package. }
  % \caption{The process of pseudo-label generation method proposed. We input the image into a pre-training model to obtain the original image label and calculate the uncertainty. We then select a reliable clip based on this uncertainty. Next, we generate pseudo-labels through random search and CycleGAN\cite{IEEEexample:Rai2018UnpairedIT} method, and calculate the uncertainty again. We select clip-level pseudo-labels based on this uncertainty. Finally, we combine the reliable and pseudo labels to create a comprehensive video classification label package.}
  \label{fig:six}
\vspace{-0.6cm}
\end{figure*}

As illustrated in \cref{fig:six}, the original-label image is generated by directly extracting the probability value from the corresponding position of the original image generated by the pre-training model. For the pseudo-label image, we utilized the pre-training module to process an input image obtained through random search and CycleGAN \cite{IEEEexample:Rai2018UnpairedIT}, and calculated the probability value of the corresponding position. The original-label image and the pseudo-label image are then combined to form a video package, which serves as the training set for the model.

Specifically, firstly, the forward network of the pre-training model calculates the probability values for each frame generated image to match the real labels. Then, using CycleGAN and random sampling techniques, each frame of static images is expanded into a package dataset, which is a combination of a sequence of continuous images, simulating a group of data frame images within a short time interval. Finally, the overall probability value of the package is obtained through weighted aggregation.

In general, when considering single-frame static images and simulated video package datasets, from the perspective of the loss function,
the weighted processing of video frames does not fundamentally differ from the processing of individual frames. Therefore, it does not affect the accuracy of data processing.
% As shown in \cref{fig:six}, we directly take the probability value of the corresponding position of the original image generated by the pre-training model as the original-label image while carrying the image obtained by random search and CycleGAN\cite{IEEEexample:Rai2018UnpairedIT} as the input of the pre-training module and calculated the probability value of the corresponding position as the pseudo-label image. Then the original-label image and the pseudo-label image are synthesized into a video package as the training set of the model.

\subsection{Optimization of Substage KD}

\paragraph{Response-based Substage}

The response-based substage, depicted in \cref{fig:short-a}, can be formulated as follows:

\begin{equation}
\ell^{Rp}_{KD}(\vartheta^{Rp}_T,\vartheta^{Rp}S)=\ell(\textRho_S,\textscy{t})+\lambda\textKappa L(\boldsymbol{\varphi}(\textscy^T) \parallel \boldsymbol{\varphi}(\textscy^S)),
\label{eq:fe-kd-loss}
\end{equation}

where $\ell(\cdot)$ is the cross-entropy loss between the student and the ground-truth labels, $\textKappa L(\cdot)$ is the KL divergence used to measure the consistency of the distribution between the teacher and the student, and $\lambda$ is a hyperparameter that balances the impact of KL divergence on the overall loss. $\textscy^T$ and $\textscy^S$ are the activation values of the teacher network as T and student network as S, $\textRho_S$ and $\textscy{t}$ are the direct output of S and the actual values of the labels, respectively.

The cross-entropy loss function $\ell(\textRho_S,\textscy_{t})$ is computed as follows:

\begin{equation}
\ell(\textRho_S,\textscy_{t}) = \sum_{i=1}^{N}\textEta(\textRho_S,\textscy_{t}),
\end{equation}

where $\textEta(\cdot)$ denotes the cross-entropy loss function, and $N$ is the capacity of the training sample space. The probability of a student classification task can be defined as $\textRho_S = [r_1, r_2, \dots, r_k, \dots, r_C]$, where $C$ denotes the number of classes. For each probability value $r_k$ in $\textRho_S$, it is computed using the softmax loss function:

\begin{equation}
r_k = \frac{exp(\textchi_{k})}{\sum^{C}_{j=1}exp(\textchi{j})}.
\end{equation}

In our case, KL divergence is used as the loss function for the classification task, which can be defined as follows:

\begin{equation}
\textKappa L = \sum^{C}_{i=1}\boldsymbol{\varphi}(y^{T}_i)\cdot\log\left(\frac{\boldsymbol{\varphi}({y^{T}_i})}{\boldsymbol{\varphi}({y^{S}i})}\right),
\end{equation}

where $y^{T}\in\textscy^T$, $y^{S}\in\textscy^S$, and $\boldsymbol{\varphi}(\cdot)$ denotes the probability distribution of activation values, which can be written as below:

\begin{equation}
\boldsymbol{\varphi}(\cdot) = \frac{\exp(\frac{y_i}{\textTau})}{\sum_{i=1}^{N}\exp(\frac{y_i}{\textTau})}
\end{equation}

Here, $\textTau$ is a hyperparameter representing the distillation temperature.

\paragraph{Feature-based Substage}

The feature-based substage, as illustrated in \cref{fig:short-c}, is a critical component of the knowledge distillation process, and it is formulated to ensure that the student network can mimic the teacher network's feature representation as accurately as possible. In this substage, the teacher-student feature mapping is accomplished by minimizing the loss function $\ell^{Fe}_{KD}(\vartheta^{Fe}_T,\vartheta^{Fe}_S)$, which is defined in \cref{eq:fe-kd-loss}.

\begin{equation}
\begin{split}
\setlength{\abovedisplayskip}{0pt}
\! \ell^{Fe}_{KD}(\vartheta^{Fe}_T,\vartheta^{Fe}_S) = \ell(\textRho_S,\textscy_{t})+\sum_{\substack{i\in\varTheta \\ k\in S_{C}}}(\alpha\textMu(\!\digamma^T_i,\!\digamma^S_i) \\
+ \beta\textMu\kappa(\!\digamma^T_{ik},\!\digamma^S_{ik})),
\setlength{\belowdisplayskip}{0pt}
\end{split}
\end{equation}

where $\varTheta$ represents the number of layer pairs that need to be mapped and distilled between the teacher and student features, and $\textMu$ uses the attention transfer \textbf{(AT)}\cite{IEEEexample:knowledgedistillation04} loss function to train the feature vector of layers in the teacher-student network. The AT method accumulates the features across the channel dimensions to achieve dimensionality reduction, allowing teachers and students to maintain the same dimension when distilling features. However, this tends to lose the distribution of features on each channel. In many cases, the feature differences between channels not only exist in the global feature information but also are closely related to the feature distribution.
% where $\varTheta$ represents the number of layer pairs that need to be mapped and distilled between the teacher and student features, and $\textMu$ uses the AT loss function in \cite{IEEEexample:knowledgedistillation04} to train the feature vector of layers in the teacher-student network. The AT method accumulates the features across the channel dimensions to achieve dimensionality reduction, allowing teachers and students to maintain the same dimension when distilling features. However, this tends to lose the distribution of features on each channel. In many cases, the feature differences between channels not only exist in the global feature information but also are closely related to the feature distribution.

To address this issue, we propose the distribution of feature dimensions \textbf{(DF)} method, which compares the distribution of feature dimensions. Based on the need to ensure the similarity of the dimension distribution of the feature layer between teachers and students, this method combines the largest K channel values in the channel dimension and their corresponding location information to the pool. It maximizes the consistency of feature dimension distribution as much as possible. $\alpha$ and $\beta$ are two hyperparameters to balance the effect of AT versus DF in the overall loss. $\textMu(\digamma^T_i,\digamma^S_i)$ and $\textMu\kappa(\digamma^T_{ik},\digamma^S_{ik})$ are defined as follows:

\begin{small} 
\begin{align}
\begin{split}
\setlength{\abovedisplayskip}{0pt}
\textMu(\digamma^T_i,\digamma^S_i) &= \parallel\frac{\digamma^T_i(\AA)}{\parallel\digamma^T_i(\AA)\parallel_2}-\frac{\digamma^S_i(\AA)}{\parallel\digamma^S_i(\AA)\parallel_2}\parallel_2 \\ 
\textMu\kappa(\digamma^T_{ik},\digamma^S_{ik}) &= \sum_{k\in S_{C}}\parallel\frac{\digamma^T_{ik}(\hat{P}\cdot\AA)}{\parallel\digamma^T_{ik}(\hat{P}\cdot\AA)\parallel_2} - \frac{\digamma^S_{ik}(\hat{P}\cdot\AA)}{\parallel\digamma^S_{ik}(\hat{P}\cdot\AA)\parallel_2}\parallel_2,
\setlength{\belowdisplayskip}{0pt}
\end{split}
\end{align}
\end{small}

where $S_C$ denotes the collection of student channels in which we want to transfer feature maps.

$\digamma^T_i(\AA) = \sum^{Ch}_{j=1}\textAlpha_{j}^2$, where Ch is the number of channels in the ith layer, $\digamma$ is a feature mapping function, which can convert the 3D features of C×H×W into corresponding 2D features and $\textAlpha_{j}$ represents the 2D feature vector on the j channel. Unlike AT, DF considers the top K channels of the maximum feature dimension of the teacher,  defined as $\digamma^T_{ik}(\hat{P}\cdot\AA) = \digamma^T_i(\hat{P_{k}}\cdot\textAlpha_{k}^2)$, where $\hat{P_{k}}$ represents the position information vector corresponding to the feature of the k channel. This vector is used to determine the optimal channel selection for the teacher, allowing for the most efficient use of the available resources.
% $\digamma^T_i(\AA) = \sum^{Ch}_{j=1}\textAlpha_{j}^2$, where Ch is the number of channels in the ith layer, $\digamma$ is a feature mapping function, which can convert the 3D features of C×H×W into corresponding 2D features and $\textAlpha_{j}$ represents the 2D feature vector on the j channel. Unlike AT, DF considers the top K channels of the maximum feature dimension of the teacher,  defined as $\digamma^T_{ik}(\hat{P}\cdot\AA) = \digamma^T_i(\hat{P_{k}}\cdot\textAlpha_{k}^2)$, where $\hat{P_{k}}$ represents the position information vector corresponding to the feature of the k channel. This vector is used to determine the optimal channel selection for the teacher, allowing for the most efficient use of the available resources.

\paragraph{Relation-based Substage}

In the relation-based substage, the goal is to measure the similarity between the internal structures of the teacher and student networks. To achieve this, we adopt the FSP matrix proposed by \cite{IEEEexample:knowledgedistillation05}, which is a matrix that measures the inner product between two feature layers in the neural network. This matrix is computed using the weights of the feature layers of the teacher and student networks and can be formulated as follows:
% \paragraph{Relation-based Substage} Relation-based substage is illustrated in \cref{fig:short-b}. We adopt the FSP matrix \cite{IEEEexample:knowledgedistillation05} to measure the similarity of the internal structure of the teacher-student network, which can be formulated as follows:

\begin{equation}
\!\ell^{Re}_{KD}(\!\vartheta^{Re}_T\!,\!\vartheta^{Re}_S)\!=\! \ell(\!\textRho_S\!,\!\textscy_{true})\! + \!\gamma\!\cdot\!\ell(\!G(W_T)\!,\!G(W_S)),
\end{equation}

where $\textRho_S$ and $\textscy_{true}$ denote the predicted and true labels of the student network, respectively. The FSP matrix $G(\cdot)$ is used to measure the similarity between the feature layers of the teacher and student networks. $\gamma$ is a hyperparameter that is used to balance the impact of FSP on the overall loss.

To compute the FSP matrix for the selected teacher-student, we denote $m$ as the selected input layer, $n$ as the selected output layer, $h$ as the height, and $w$ as the width. The FSP matrix can be computed as follows:
% where the FSP matrix $G(\cdot)$ denotes the inner product between two feature layers in the neural network, $W_T$ and $W_S$ represent the weights of the feature layers of the teacher and student networks, and $\gamma$ is a hyperparameter used to balance the impact of FSP on the overall loss. Let $m$ denote the selected input layer, $n$ the selected output layer, $h$ the height, and $w$ the width. The FSP matrix for the selected teacher-student is computed as follows:

\begin{equation}
G_(m,n)(\cdot) = \sum_{i=1}^{h}\sum_{j=1}^{w}\frac{W_{(\cdot)m} \times W_{(\cdot)n}}{h \times w}.
\end{equation}

Here, $W_{(\cdot)m}$ and $W_{(\cdot)n}$ are the weight matrices of the input and output layers of the selected teacher-student, respectively. The FSP matrix is normalized by the product of the height and width of the feature map to ensure that the values of the matrix are between 0 and 1. This normalization helps to reduce the impact of the size of the feature map on the FSP matrix.

The relation-based substage helps the student network to mimic the internal structure of the teacher network by measuring the similarity between their feature layers. The FSP matrix is a powerful tool for measuring this similarity and can help to improve the performance of the student network.

\subsection{Combination Stage}

\begin{figure*}[h]
  \begin{subfigure}{.49\linewidth}
	\centering
    %\fbox{\rule{0pt}{2in} \rule{.9\linewidth}{0pt}}
    \includegraphics[width=1\linewidth]{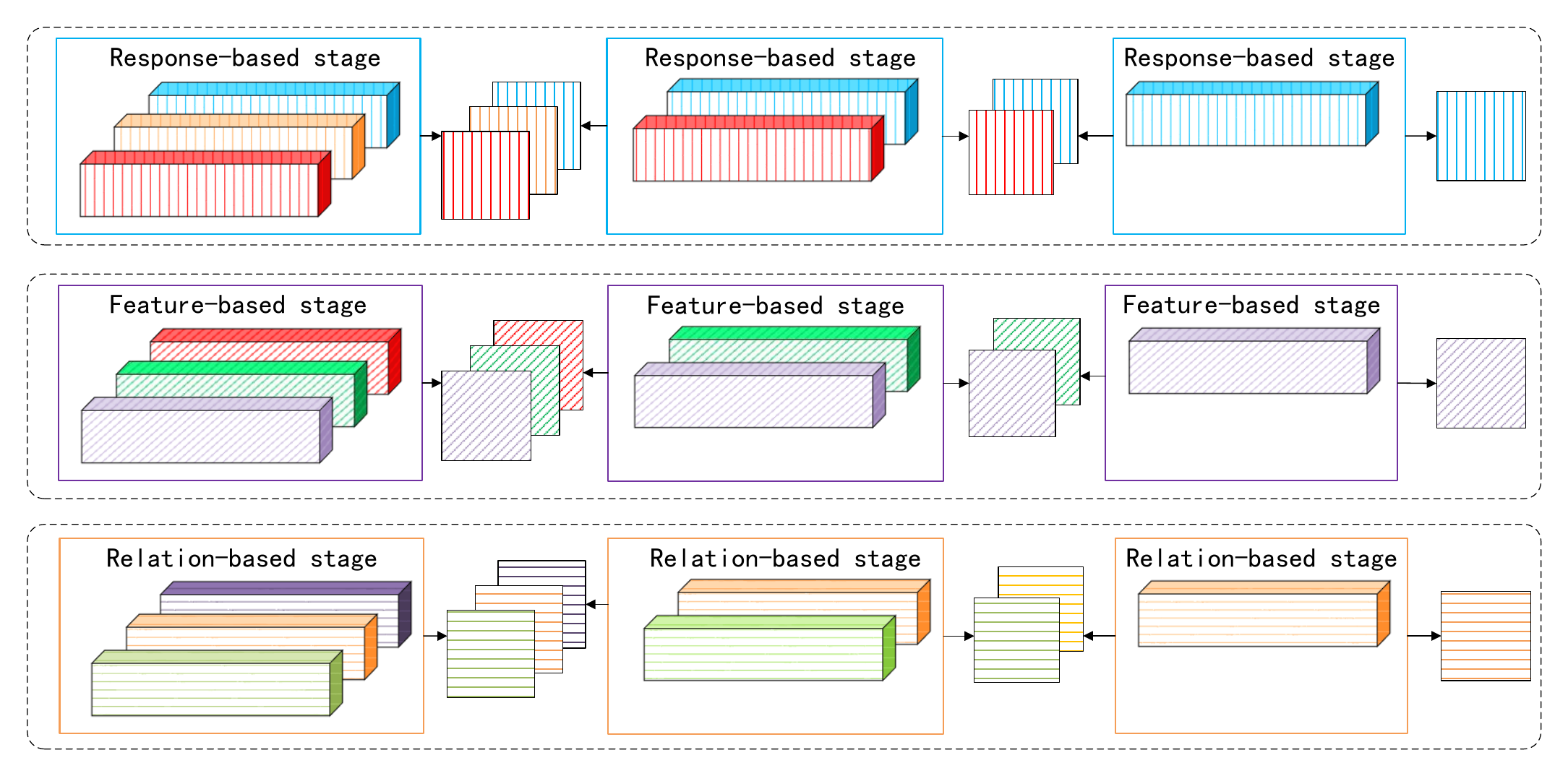}
	\caption{SKD stage.}
    \label{fig:forth-a}
  \end{subfigure}
  \hfill
  \begin{subfigure}{.53\linewidth}
    \centering
    %\fbox{\rule{0pt}{2in} \rule{.9\linewidth}{0pt}}
    \includegraphics[width=1\linewidth]{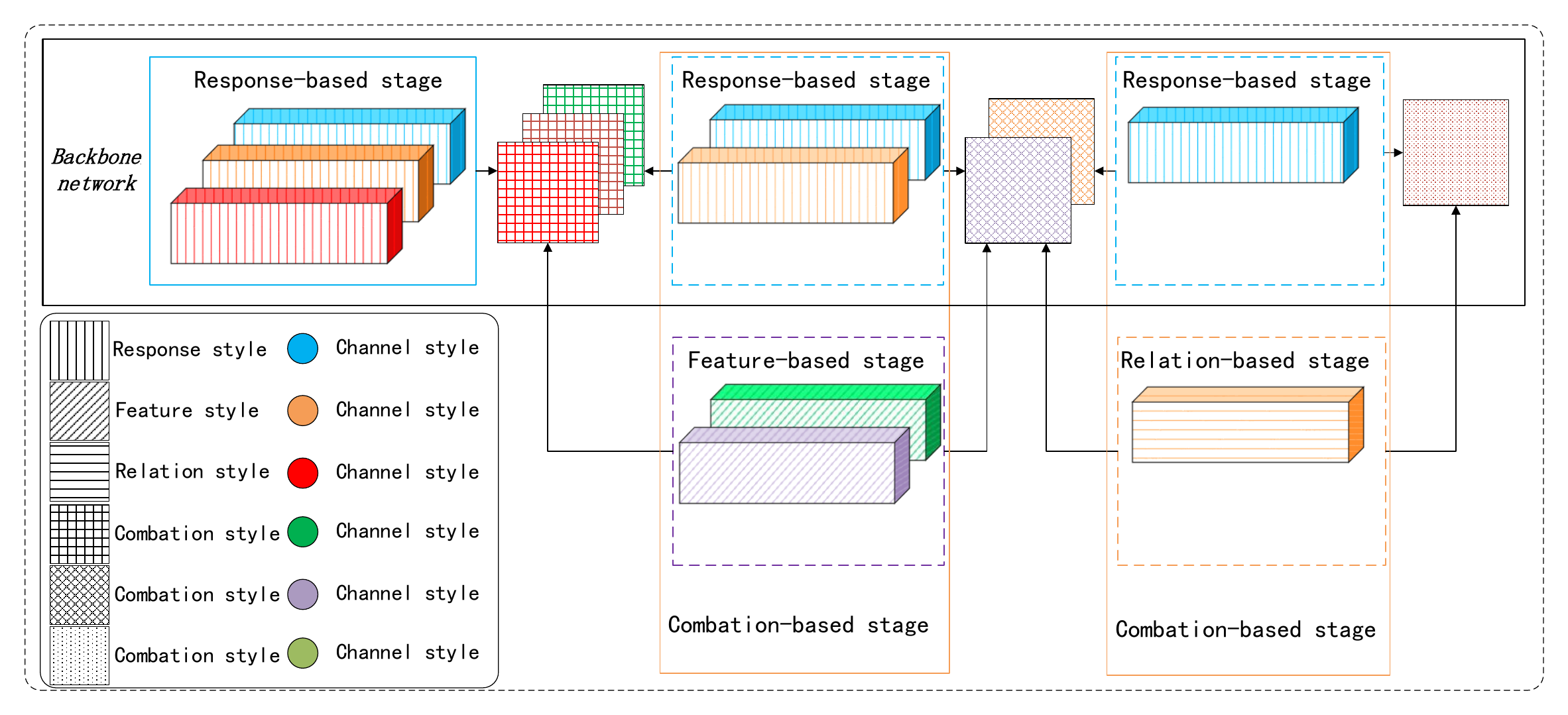}
    \caption{RSKD stage.}
    \label{fig:forth-b}
  \end{subfigure}
  \caption{Illustration of the processes of SKD and RSKD knowledge distillation, with divergent substage styles represented by distinct lines and assorted channel attribute styles by distinct hues. In SKD (a), the knowledge distillation process consists of three independent substages, while in RSKD (b), it includes the combination-based stage and the response-based KD substage implemented in the cascaded backbone network.}
\vspace{-0.6cm}
  \label{fig:forth}
\end{figure*}

\renewcommand{\algorithmicrequire}{ \textbf{Input:}} %Use Input in the format of Algorithm
\renewcommand{\algorithmicensure}{ \textbf{Output:}} %UseOutput in the format of Algorithm
\begin{algorithm}[htb]
\caption{Transfer the distilled knowledge with SKD}
\label{alg:A}
\begin{algorithmic}[1] %这个1 表示每一行都显示数字
\REQUIRE ~~\\ %算法的输入参数：Input
    Teacher pre-training weight of each substage $\boldsymbol{\vartheta}^{Rp}_{\boldsymbol{T_1}}$, $\boldsymbol{\vartheta}^{Fe}_{\boldsymbol{T_1}}$, $\boldsymbol{\vartheta}^{Re}_{\boldsymbol{T_1}}$

	Students randomly initialized weight of each substage $\boldsymbol{\vartheta}^{Rp}_{\boldsymbol{S_1}}$, $\boldsymbol{\vartheta}^{Fe}_{\boldsymbol{S_1}}$, $\boldsymbol{\vartheta}^{Re}_{\boldsymbol{S_1}}$

    Number of cascading layers N
\ENSURE ~~\\ %算法的输出：Output
    Student training weight $\boldsymbol{\vartheta}^{Integration}_{\boldsymbol{S_{final}}}$
    
\textbf{Stage 1:} Initial stage distillation
	\STATE {$\boldsymbol{\vartheta}^{Rp}_{\boldsymbol{S_1}}$ = arg min($\ell^{Rp}_{KD}$($\boldsymbol{\vartheta}^{Rp}_{\boldsymbol{T_1}}$ , $\boldsymbol{\vartheta}^{Rp}_{\boldsymbol{S_1}}$))}
	\STATE {$\boldsymbol{\vartheta}^{Fe}_{\boldsymbol{S_1}}$ = arg min($\ell^{Fe}_{KD}$($\boldsymbol{\vartheta}^{Fe}_{\boldsymbol{T_1}}$ , $\boldsymbol{\vartheta}^{Fe}_{\boldsymbol{S_1}}$))}
	\STATE {$\boldsymbol{\vartheta}^{Re}_{\boldsymbol{S_1}}$ = arg min($\ell^{Re}_{KD}$($\boldsymbol{\vartheta}^{Re}_{\boldsymbol{T_1}}$ , $\boldsymbol{\vartheta}^{Re}_{\boldsymbol{S_1}}$))}

\textbf{Stage 2:} Cascade stage distillation
    	\FOR{$i=1;i \le N;$}
		\STATE {$\boldsymbol{\vartheta}^{Rp}_{S_{i+1}}$ = arg min($\ell^{Rp}_{KD}$($\boldsymbol{\vartheta}^{Rp}_{\boldsymbol{S_i}}$ , $\boldsymbol{\vartheta}^{Rp}_{\boldsymbol{S_{i+1}}}$))}	
		\STATE {$\boldsymbol{\vartheta}^{Fe}_{\boldsymbol{S_{i+1}}}$ = arg min($\ell^{Fe}_{KD}$($\boldsymbol{\vartheta}^{Fe}_{\boldsymbol{S_i}}$ , $\boldsymbol{\vartheta}^{Fe}_{\boldsymbol{S_{i+1}}}$))}
		\STATE {$\boldsymbol{\vartheta}^{Re}_{\boldsymbol{S_{i+1}}}$ = arg min($\ell^{Re}_{KD}$($\boldsymbol{\vartheta}^{Re}_{\boldsymbol{S_i}}$ , $\boldsymbol{\vartheta}^{Re}_{\boldsymbol{S_{i+1}}}$))}
		\ENDFOR

\textbf{Stage 3:} Integration stage
\STATE {$\boldsymbol{\vartheta}^{Integration}_{\boldsymbol{S_{final}}}$ = avg($\boldsymbol{\vartheta}^{Rp}_{\boldsymbol{S_{N+1}}}$,$\boldsymbol{\vartheta}^{Fe}_{\boldsymbol{S_{N+1}}}$,$\boldsymbol{\vartheta}^{Re}_{\boldsymbol{S_{N+1}}}$)}
\end{algorithmic}
\end{algorithm}
\begin{algorithm}[htb]
\caption{Transfer the distilled knowledge with RSKD}
\label{alg:B}
\begin{algorithmic}[1] %这个1 表示每一行都显示数字
\REQUIRE ~~\\ %算法的输入参数：Input
    Teacher pre-training weight of each substage $\boldsymbol{\vartheta}^{Rp}_{\boldsymbol{T_1}}$, $\boldsymbol{\vartheta}^{Fe}_{\boldsymbol{T_1}}$, $\boldsymbol{\vartheta}^{Re}_{\boldsymbol{T_1}}$

	Students randomly initialized weight of each combination-stage $\boldsymbol{\vartheta}^{Rp-CS}_{\boldsymbol{S_1}}$, $\boldsymbol{\vartheta}^{Fe-CS}_{\boldsymbol{S_1}}$, $\boldsymbol{\vartheta}^{Re-CS}_{\boldsymbol{S_1}}$

    Number of cascading layers N
\ENSURE ~~\\ %算法的输出：Output
    Student training weight $\boldsymbol{\vartheta}^{Total}_{\boldsymbol{S_{final}}}$

\textbf{Stage 1:} Initial stage distillation
	\STATE {$\boldsymbol{\vartheta}^{Rp-CS}_{\boldsymbol{S_1}}$ = arg min($\ell^{Rp-CS}_{KD}$($\boldsymbol{\vartheta}^{Rp}_{\boldsymbol{T_1}}$ , $\boldsymbol{\vartheta}^{Rp-CS}_{\boldsymbol{S_1}}$))}
	\STATE {$\boldsymbol{\vartheta}^{Fe-CS}_{\boldsymbol{S_1}}$ = arg min($\ell^{Fe-CS}_{KD}$($\boldsymbol{\vartheta}^{Fe}_{\boldsymbol{T_1}}$ , $\boldsymbol{\vartheta}^{Fe-CS}_{\boldsymbol{S_1}}$))}
	\STATE {$\boldsymbol{\vartheta}^{Re-CS}_{\boldsymbol{S_1}}$ = arg min($\ell^{Re-CS}_{KD}$($\boldsymbol{\vartheta}^{Re}_{\boldsymbol{T_1}}$ , $\boldsymbol{\vartheta}^{Re-CS}_{\boldsymbol{S_1}}$))}

\textbf{Stage 2:} Cascade stage distillation
    	\FOR{$i=1;i \le N;$}
		\STATE {$\boldsymbol{\vartheta}^{Rp-CS}_{S_{i+1}}$ = arg min($\ell^{Rp-CS}_{KD}$($\boldsymbol{\vartheta}^{Rp-CS}_{\boldsymbol{S_i}}$ , $\boldsymbol{\vartheta}^{Rp-CS}_{\boldsymbol{S_{i+1}}}$))}
		\STATE {$\boldsymbol{\vartheta}^{Fe-CS}_{\boldsymbol{S_{i+1}}}$ = arg min($\ell^{Fe-CS}_{KD}$($\boldsymbol{\vartheta}^{Fe-CS}_{\boldsymbol{S_i}}$ , $\boldsymbol{\vartheta}^{Fe-CS}_{\boldsymbol{S_{i+1}}}$))}
		\STATE {$\boldsymbol{\vartheta}^{Re-CS}_{\boldsymbol{S_{i+1}}}$ = arg min($\ell^{Re-CS}_{KD}$($\boldsymbol{\vartheta}^{Re-CS}_{\boldsymbol{S_i}}$ , $\boldsymbol{\vartheta}^{Re-CS}_{\boldsymbol{S_{i+1}}}$))}
		\ENDFOR

\textbf{Stage 3:} Total stage distillation
\STATE {$\boldsymbol{\vartheta}^{Total}_{\boldsymbol{S_{final}}}$ = avg($\boldsymbol{\vartheta}^{Rp-CS}_{S_{N+1}}$,$\boldsymbol{\vartheta}^{Fe-CS}_{\boldsymbol{S_{N+1}}}$,$\boldsymbol{\vartheta}^{Re-CS}_{\boldsymbol{S_{N+1}}}$)}
\end{algorithmic}
\end{algorithm}

In the proposed improved \textbf{(CS)} design method, as shown in \cref{fig:forth-b}, auxiliary branch substages are added in parallel for distillation, with a response-based substage serving as the backbone. Different combination stages are then composed of substages from different branches. This method introduces diversity in the substages used for distillation, which helps students learn more diverse and essential knowledge from teachers, and thus improves overall accuracy.

Although the diversity introduced may not be as accurate as the corresponding substages based on the response, it indirectly facilitates the learning of deep semantic features of distillation. This approach changes the style of substages of branches and the style of channels simultaneously, which enhances the generalization ability of students. This is particularly useful when the same substage is used for each distillation branch, which results in high style similarity between adjacent substages within the same branch, hindering the learning process of adjacent substages and diminishing the generalization ability of students, as shown in \cref{fig:forth-a}.

It is worth noting that the combination of CS branches and the backbone network structure significantly affects the final KD. As ensemble models have demonstrated, the diversity of classifiers plays an essential role in boosting performance \cite{IEEEexample:MOD}. Similarly, the diversity introduced in the substages used for distillation can significantly improve the performance of KD. The improved CS design method achieves this by introducing diversity in the substages used for distillation.

\subsection{SKD and RSKD}

The proposed SKD framework is based on the observation that different stages of KD have a phased impact on the final learning process. The training process of SKD mainly consists of three stages: response-based, feature-based, and relation-based. Each stage has a distinct purpose, allowing for a more comprehensive learning experience.
% Our proposed SKD framework is based on the fact that different stages of KD have a phased impact on the final learning process. The training process of SKD mainly consists of three stages: response-based, feature-based, and relation-based. Each stage has a distinct purpose, allowing for a more comprehensive learning experience.

The response-based stage distills the teacher-student final output result, which is similar to the final result in the human learning process. The feature-based stage distills the specific layers that the teacher-student selects to distill, making them as similar as possible, akin to the shorter cycle in the human learning process. Lastly, the relation-based stage distills the similarity between specific layers at different scales, replicating the phased outcomes learned in the long human learning cycle. This multi-stage approach enables the student to learn both high-level and low-level knowledge from the teacher network.
% Response-based distills the teacher-student final output result, similar to the final result in the human learning process. Feature-based distills the specific layers that the teacher-student selects to distill, making them as similar as possible, akin to the shorter cycle in the human learning process. Lastly, relation-based distills the similarity between specific layers at different scales, replicating the phased outcomes learned in the long human learning cycle.

The SKD training process is outlined in Algorithm \ref{alg:A}. By leveraging the distinct advantages of each stage, our SKD framework provides a comprehensive and effective learning experience. However, integrating the output results of the three stages would lead to memory consumption and delay in the deployment model.
% The SKD training process is outlined in Algorithm \ref{alg:A}. By leveraging the distinct advantages of each stage, our SKD framework provides a comprehensive and effective learning experience. As a result of SKD, integrating the output results of three stages would lead to memory consumption and delay of the deployment model.

To address this, the proposed RSKD method forgoes the idea of an ensemble and instead uses a combination stage structure to combine the output results of different branch stages in the distillation process. This approach improves the model's generalization performance and reduces memory consumption and inference delay, achieving a balance between precision and efficiency, making it more conducive to model deployment in small embedded devices. As shown in \cref{fig:forth-b}, the RSKD knowledge distillation process consists of a combination stage where different types of substage designs are used for each branch, with the backbone network using a response-based substage. The cascaded backbone network uses the response-based KD substage, and other auxiliary branch substages are added in parallel for distillation. \Cref{alg:B} details the RSKD training process.
% To address this, RSKD forgoes the idea of an ensemble and instead uses a CS structure to assist the output results of different branch stages in the distillation. That improves the model's generalization performance and reduces memory consumption and inference delay, achieving a balance between precision and efficiency, making it more conducive to model deployment in small embedded devices. \cref{alg:B} details the RSKD training process.

%-------------------------------------------------------------------------
\section{Experiments}

\subsection{Experimental Setup}

\paragraph{Datasets}

We perform experiments on two public image classification benchmarks: CIFAR-100 and ImageNet. To synthesize pseudo-labels for the video classification images, we follow the method shown in \cref{fig:six}, which allows us to leverage the rich information in videos to improve image classification performance. We also used a real video action recognition dataset: UCF101 \cite{IEEEexample:Soomro2012UCF101AD}.

CIFAR-100 is a widely used image classification dataset that contains 32x32 images of 100 categories. The dataset comprises 50,000 training images, with 500 images for each class, and 10,000 test images, with 100 images for each class.

ImageNet is a well-known large-scale image classification dataset that consists of 1.2 million training images from 1,000 classes and 50,000 validation images. Unlike CIFAR-100, ImageNet better represents the real-world diversity and complexity of images.

UCF101 is a widely-used dataset for video action recognition, comprising around 13,000 video clips sourced from YouTube. It encompasses 101 distinct action categories, with each video clip having an average duration of approximately 7 seconds and a fixed resolution of 320x240 pixels.

\paragraph{Network Structure}

We evaluate our method on two types of network structures: homogeneous and heterogeneous. Homogeneous networks refer to the network structures of a teacher and student network with the same architecture, allowing us to measure the performance gain of distillation in a like-for-like scenario. We consider three architectures for homogeneous networks: ResNet \cite{IEEEexample:classfy04}, Wide ResNet \cite{IEEEexample:classfy05}, and VGG \cite{IEEEexample:classfy01}. For heterogeneous networks, we test our method on networks with different architectures for the teacher and student networks, which allows us to evaluate the generalization ability of our method across architectures. We consider four architectures for heterogeneous networks: ResNet, Wide ResNet, ShuffleNet \cite{IEEEexample:modelCompact03}, and MobileNet \cite{IEEEexample:modelCompact01}.
% We divide network structures into two types: homogeneous networks and heterogeneous networks. Homogeneous network refers to the network structure of a teacher network and student network with the same nature, and its purpose is to verify the performance of our method in the same type of networks, mainly including ResNet\cite{IEEEexample:classfy04}, Wide ResNet\cite{IEEEexample:classfy05} and VGG\cite{IEEEexample:classfy01}. A heterogeneous network means that the teacher network and student network have different network structures, and its purpose is to verify the performance and generalization ability of our method in networks with different structures, mainly including ResNet\cite{IEEEexample:classfy04}, Wide ResNet\cite{IEEEexample:classfy05}, ShuffleNet\cite{IEEEexample:modelCompact03} and MobileNet\cite{IEEEexample:modelCompact01}.

\paragraph{Implementation Details}

\begin{table}[h]
\caption{The weight factor $\beta$ for different networks.} \label{t0}
\centering
\resizebox{1.\linewidth}{!}{
\begin{tabular}{c|cccccc}
\multirow{1}{*}{}
& FITNET\cite{IEEEexample:knowledgedistillation02} & AT\cite{IEEEexample:knowledgedistillation04} & AB\cite{IEEEexample:AB} & FSP\cite{IEEEexample:knowledgedistillation05} & CRD\cite{IEEEexample:CRD} & DKD\cite{IEEEexample:DKD}\\
\hline
\multirow{1}{*}{$\beta$}
& 100 & 1000 & 0 & 0 & 0.8 & 8\\
\end{tabular}}
\end{table}

To verify our method, we compared our proposed SKD and RSKD with KD\cite{IEEEexample:knowledgedistillation01}, FITNET\cite{IEEEexample:knowledgedistillation02}, AT\cite{IEEEexample:knowledgedistillation04}, AB\cite{IEEEexample:AB}, FSP\cite{IEEEexample:knowledgedistillation05}, CRD\cite{IEEEexample:CRD} and DKD\cite{IEEEexample:DKD} methods. To ensure a fair comparison, we implemented the algorithms of other methods based on their papers and codes.
% In order to verify our method, we compare our proposed SKD and RSKD with KD\cite{IEEEexample:knowledgedistillation01}, FITNET\cite{IEEEexample:knowledgedistillation02}, AT\cite{IEEEexample:knowledgedistillation04}, AB\cite{IEEEexample:AB}, FSP\cite{IEEEexample:knowledgedistillation05}, CRD\cite{IEEEexample:CRD} and DKD\cite{IEEEexample:DKD} methods. To prevent other factors from interfering with the fairness of the comparison, we reproduce the algorithms of other methods based on their papers and codes. For CIFAR-100 datasets, we filled 4 pixels on each side to expand the original image size to 40×40 pixels. Then we used a randomly cropped 32×32 pixel image for training while retaining the original 32×32 pixel image for testing.

For CIFAR-100 datasets, we expanded the original 32x32 images to 40x40 by filling 4 pixels on each side. We used randomly cropped 32x32 pixel images for training and kept the original 32x32 images for testing. We used pre-trained teachers as the initial teacher network. To maintain consistency, we set the weight factor $\beta$ that balances the actual loss with the distillation loss to the optimal value specified in the original text if available. Otherwise, we used the value obtained by grid search. The weight factor $\beta$ of different networks is shown in \cref{t0}.
% The Initial teacher Network uses pre-trained teachers. Some important hyperparameters need to be consistent. For the weight factor $\beta$ that balances the actual loss with the distillation loss, we directly use the optimal value in the original text if it is specified in the original text. Otherwise, we use the value obtained by grid search. The weight factor $\beta$ of different networks is shown in \cref{t0}.

For the distillation of AB\cite{IEEEexample:AB} and FSP\cite{IEEEexample:knowledgedistillation05}, we used a separate pre-training phase, so $\beta$ was set to 0. The $\alpha$ of DKD\cite{IEEEexample:DKD} network was set to 1. We set the temperature parameter T = 4 and $\alpha$ = 0.9 for KD following \cite{IEEEexample:knowledgedistillation01}. All methods in our experiment were evaluated using SGD. Our SKD and RSKD algorithms used three substage cascades per process by default, and we set $\lambda$ to 0.9, $\alpha$ to 200, $\beta$ to 300, and $\gamma$ to 0.9.
% The distillation of AB\cite{IEEEexample:AB} and FSP\cite{IEEEexample:knowledgedistillation05} take place in a separate pre-training phase, so $\beta$ is set to 0. $\alpha$ of DKD\cite{IEEEexample:DKD} network is set to 1. We set the temperature parameter T = 4 and $\alpha$ = 0.9 for KD following \cite{IEEEexample:knowledgedistillation01}. All the methods in our experiment are evaluated using SGD. Our SKD and RSKD algorithms default to use three substage cascades per process, and we set $\lambda$ to 0.9, $\alpha$ to 200, $\beta$ to 300, and $\gamma$ to 0.9.

For CIFAR-100, the initial learning rate of each stage cascade was 0.01 for MobileNetV2\cite{IEEEexample:modelCompact02}, ShuffleNetV1\cite{IEEEexample:modelCompact03}, and ShuffleNetV2\cite{IEEEexample:Shufflenetv2}, and 0.05 for other models. The learning rate decayed by a factor of 0.1 every 30 epochs after the first 120 epochs until the last 150 epochs, and the batch size was set to 128.
% For CIFAR-100, each stage cascade has an initial learning rate of 0.01 for MobileNetV2\cite{IEEEexample:modelCompact02}, ShuffleNetV1\cite{IEEEexample:modelCompact03}, and ShuffleNetV2\cite{IEEEexample:Shufflenetv2}, while the learning rate is 0.05 for the other models. The learning rate decays 0.1 every 30 epochs after the first 120 epochs until the last 150 epochs, and the batch size is 128 for CIFAR-100.

For ImageNet, we followed the standard PyTorch\cite{IEEEexample:pytorch} practice, with each stage cascade having an initial learning rate of 0.2, a weight decay of 0.001 using the MSRA initialization technique\cite{IEEEexample:MSRA}, and a batch size of 256.
% For ImageNet, we follow the standard PyTorch\cite{IEEEexample:pytorch} practice, and each stage cascade has an initial learning rate of 0.2, while different from the weight decay strategy adopted by CIFAR-100, with a weight decay of 0.001 using the MSRA initialization technique\cite{IEEEexample:MSRA}, and the batch size is 256. The PyTorch framework implements all training and validation processes on an NVIDIA Tesla P100 industrial GPU server.

For UCF101, the pretrained model was trained from scratch using randomly initialized weights. Initially, it was trained for 200 epochs using SGD with a learning rate of 0.01 and weight decay of 0.0001. Subsequently, it was trained for an additional 100 epochs using Adam optimizer with a learning rate of 0.0002 and weight decay of 0.05. The batch size was set to 16 clips per GPU.

All training and validation processes were implemented using the PyTorch framework on an NVIDIA Tesla P100 industrial GPU server.
% We used \cref{t11} and \cref{t12} to explain the architecture of the cascading models used in the experiment for the same and different network structures.

%-------------------------------------------------------------------------

\subsection{Details about RSKD}

\paragraph{Formulation} 

The final distillation result of our formalized RSKD is given as follows:

\begin{equation}
\!\ell^{final}_{KD}\!=\!\ell(\textRho_S,\textscy_{t})\!+\!\eta\cdot\!\ell^{Rp-CS}_{KD} +\!\xi\cdot\!\ell^{Fe-CS}_{KD}\! +\!\tau\cdot\!\ell^{Re-CS}_{KD},
\end{equation}

where $\ell(\textRho_S,\textscy_{t})$ represents the standard knowledge distillation loss, and $\eta$, $\xi$, and $\tau$ are hyperparameters that control the importance of the consistency losses. The $^{Rp-CS}$, $^{Fe-CS}$, and $^{Re-CS}$ superscripts refer to the response-based, feature-based, and response-feature-based consistency losses, respectively.

We believe that the effectiveness of RSKD for knowledge transfer may depend on the complexity of the datasets involved. Similar to the learning process of humans, when the knowledge to be mastered is elementary, the learning process is short and resembles the response-based substage. On the other hand, when the knowledge to be learned is complex, the learning process is long and resembles the feature-based and response-based substages. Therefore, inappropriate ratios of $\eta$, $\xi$, and $\tau$ may compromise the accuracy of the student model's predictions.
% We believe that the importance of RSKD in knowledge transfer may be related to the complexity of data sets. Intuitively, it is similar to the learning process of human beings. Suppose the knowledge to be mastered is elementary. In that case, the learning process will be short and similar to the response-based substage, and if the knowledge to be learned is very complex, the learning process will be long and similar to the feature-based and response-based substages. Therefore, inappropriate ratios of $\eta$, $\xi$, and $\tau$ may compromise the correctness of students' predictions.

%-------------------------------------------------------------------------
\subsection{Comparison of Training Accuracy}

\paragraph{Video Classification on CIFAR-100}

The verification accuracy of top-1 is shown in \cref{t1} and \cref{t2}, where we evaluate RSKD on CIFAR-100. In the absence of an ensemble, \cref{t1} presents the experimental results of teacher-student networks with the same network structure style, while \cref{t2} shows the results of teacher-student networks with different network structure styles.
% The verification accuracy of top-1 is shown in \cref{t1} and \cref{t2}. We test RSKD on CIFAR-100. In the absence of an ensemble, \cref{t1} presents the experimental results of teacher-student networks with the same network structure style. \cref{t2} shows the experimental results of teacher-student networks with different network structure styles.

It is worth noting that while RSKD does not always achieve the optimal value in some teacher-student combinations when the teacher and student network styles are the same, it outperforms other distillation methods in experiments of different network architectures. This phenomenon can be explained by the fact that the small size of CIFAR-100 data may not fully demonstrate the advantages of stage learning in the same style of network architecture. Further discussion on this topic can be found in \ref{t3}, \ref{t4}, \ref{t5}, and \ref{t6}.
% It is worth noting that, compared with other distillation methods, RSKD does not always achieve the optimal value in some teacher-student combinations when the teacher and student network styles are the same. We think that the main reason for this phenomenon is that the small size of CIFAR-100 data cannot fully demonstrate the advantages of stage learning in the same style of network architecture. \cref{t3} and \cref{t4} can be used to discuss this further.

The experimental results confirm that the CS structure in RSKD can uncover the hidden knowledge in different network architectures, leading to better distillation performance.
% At the same time, in the experiments of different network architectures, RSKD outperforms other methods. The CS structure can uncover the hidden knowledge in different network architectures, which helps to obtain a better distillation effect.

\begin{table*}
\caption{Results of student networks on CIFAR-100 (\%), where teachers and students transfer across the same architectures. The table includes results obtained using a cascading strategy with three layers, denoted by $\ast$. The reported results are averaged over more than five runs.} \label{t1}
\centering
\resizebox{.87\linewidth}{!}{
\begin{tabular}{ccccccc}
\bottomrule
\multirow{5}{*}{Distillation Stage}
& Teacher Network & VGG13 & WideResNet-40-2 & ResNet56 & ResNet32×4 & ResNet110\\
& Student Network & VGG8 & WideResNet-16-2 & ResNet20 & ResNet8×4 & ResNet32\\
\cmidrule{2-7}
& Teacher & 74.64 & 75.61 & 72.34 & 79.42 & 74.31\\
& Student & 70.36 & 73.26 & 69.06 & 72.50 & 71.14\\
\bottomrule
\multirow{3}{*}{Response Stage}
& $\ast$KD\cite{IEEEexample:knowledgedistillation01} & 73.98 & 75.92 & 71.26 & 73.69 & 73.58\\
& $\ast$AB\cite{IEEEexample:AB} & 71.54 & 72.90 & 69.71 & 73.52 & 71.26\\
& $\ast$DKD\cite{IEEEexample:DKD} & \textbf{74.61} & 76.29 & \textbf{71.96} & 76.33 & 74.13\\
\bottomrule
\multirow{3}{*}{Feature Stage}
& $\ast$FITNET\cite{IEEEexample:knowledgedistillation02} & 71.32 & 73.68 & 69.53 & 74.16 & 71.36\\
& AT\cite{IEEEexample:knowledgedistillation04} & 71.63 & 74.31 & 70.65 & 73.97 & 72.54\\
& $\ast$CRD\cite{IEEEexample:CRD} & 73.98 & 75.67 & 71.32 & 75.71 & 73.69\\
\bottomrule
\multirow{1}{*}{Relation Stage}
& $\ast$FSP\cite{IEEEexample:knowledgedistillation05} & 70.63 & 73.10 & 70.04 & 72.82 & 71.95\\
\bottomrule
\multirow{1}{*}{Combination Stage}
& \textbf{RSKD} & 74.59 & \textbf{76.32} & 71.82 & \textbf{76.53} & \textbf{74.21}\\
\bottomrule
\end{tabular}}
\end{table*}

\begin{table*}
\caption{Results of student networks on CIFAR-100 (\%), where teachers and students transfer across different architectures. The table includes results obtained using a cascading strategy with three layers, denoted by $\ast$. The $\dagger$ symbol denotes the removal of the relation-based stage. The reported results are averaged over more than five runs.} \label{t2}
\centering
\resizebox{0.87\linewidth}{!}{
\begin{tabular}{cccccc}
\bottomrule
\multirow{5}{*}{Distillation Stage}
& Teacher Network & ResNet50 & ResNet32×4 & ResNet32×4 & WideResNet-40-2\\
& Student Network & MobileNetV2 & ShuffleNetV1 & ShuffleNetV2 & ShuffleNetV1\\
\cmidrule{2-6}
& Teacher & 79.34 & 79.42 & 79.42 & 75.61\\
& Student & 64.60 & 70.50 & 71.82 & 70.50\\
\bottomrule
\multirow{3}{*}{Response Stage}
& $\ast$KD\cite{IEEEexample:knowledgedistillation01} & 68.51 & 75.11 & 75.35 & 75.94\\
& $\ast$AB\cite{IEEEexample:AB} & 67.76 & 73.79 & 74.83 & 73.69\\
& $\ast$DKD\cite{IEEEexample:DKD} & 70.37 & 76.48 & 77.09 & 76.72\\
\bottomrule
\multirow{3}{*}{Feature Stage}
& $\ast$FITNET\cite{IEEEexample:knowledgedistillation02} & 63.19 & 73.70 & 73.61 & 73.91\\
& AT\cite{IEEEexample:knowledgedistillation04} & 59.02 & 71.92 & 72.91 & 73.46\\
& $\ast$CRD\cite{IEEEexample:CRD} & 69.32 & 75.27 & 75.76 & 76.25\\
\bottomrule
\multirow{1}{*}{Combination Stage}
& \textbf{RSKD$\dagger$} & \textbf{70.55} & \textbf{76.57} & \textbf{77.11} & \textbf{76.85}\\
\bottomrule
\end{tabular}}
\end{table*}

\paragraph{Video Classification on ImageNet}

We evaluate our SKD, RSKD, and $\bullet$RSKD on ImageNet, and present the verification accuracy of top-1 and top-5 in \cref{t3} and \cref{t4}, respectively. \cref{t3} contains the results of teachers and students having the same network architecture, while \cref{t4} shows the results for teachers and students from different network architectures. To ensure experimental fairness, we employ ensemble and cascading strategies in all baseline models. Specifically, we use a three-level cascading and three-branch ensemble configuration as the default setting, allowing for a more accurate performance comparison.
% We discuss the experimental results of ImageNet to test our SKD and RSKD. The verification accuracy of top-1 and top-5 is shown in \cref{t3} and \cref{t4}, respectively. \cref{t3} contains the results of teachers and students having the same network architecture, while \cref{t4} shows the results for teachers and students from different network architectures. Ensemble and cascading strategies are employed in all baseline models compared for experimental fairness. The default setting is a three-level cascading and three-branch ensemble configuration, which ensures that all models are evaluated on an even playing field, allowing for more accurate performance comparison.

The experimental results confirm our theoretical hypothesis that our SKD, RSKD, and $\bullet$RSKD achieve significant improvements on ImageNet with the increase of the data set. RSKD improves the accuracy of top-1 and top-5 by about 1\% when compared with networks with the same teacher-student network style, while $\bullet$RSKD improves the accuracy of top-1 and top-5 by about 1\%-1.5\%. For networks with different teacher-student network styles, RSKD improves the accuracy of top-1 and top-5 by about 1\%-1.5\%, while $\bullet$RSKD improves the accuracy of top-1 and top-5 by about 1.5\%-2\%. Our methods outperform the most state-of-the-art distillation methods on ImageNet.

\paragraph{Video Classification on UCF101}

We assessed the performance of our SKD, RSKD, and $\bullet$ RSKD techniques on the UCF101 dataset, and presented the validation accuracies for top-1 and top-5 in \ref{t5} and \ref{t6} respectively. \ref{t5} encompasses the outcomes for both teachers and students with identical network architectures, while \ref{t6} showcases the results obtained from teachers and students with distinct network architectures. To ensure the fairness of our experiments, we employed ensemble and cascading strategies across all baseline models. Specifically, we employed a default setup of a three-stage cascading and three-branch ensemble configuration, allowing for more precise performance comparisons.
% The experimental results confirm our theoretical hypothesis that our SKD, RSKD, and $\bullet$RSKD have achieved significant improvements on ImageNet with the increase of the data set, all of which are superior to the results obtained by the most state-of-the-art distillation methods. On the ImageNet dataset, RSKD improves the accuracy of top-1 and top-5 by about 1\%, while $\bullet$RSKD improves the accuracy of top-1 and top-5 by about 1\%-1.5\% when compared with networks with the same teacher-student network style. For networks with different teacher-student network styles, RSKD improves the accuracy of top-1 and top-5 by about 1\%-1.5\%, while $\bullet$RSKD improves the accuracy of top-1 and top-5 by about 1.5\%-2\%.

The experimental results confirm our theoretical hypothesis that with the increase of real video data, our SKD, RSKD, and $\bullet$ RSKD techniques achieve significant improvements on UCF101. In comparison to networks with the same teacher-student network architecture, RSKD enhances the top-1 and top-5 accuracy by approximately 2\%, while $\bullet$ RSKD achieves an improvement of about 2\% - 3\% in both top-1 and top-5 accuracy. For networks with different teacher-student network architectures, RSKD improves the top-1 and top-5 accuracy by approximately 2\% - 2.5\%, whereas $\bullet$ RSKD yields an enhancement of around 4\% - 5\% in top-1 and top-5 accuracy.

\begin{table*}
\caption{Top-1 and Top-5 accuracy (\%) of the student network ResNet-18 on ImageNet with teacher and student transfer across the same architectures. ResNet-34 is used as the teacher network and ResNet-18 as the student network. The $\bullet$ symbol indicates the ensemble strategy using three branches, and the $\ast$ symbol indicates the cascading strategy using three layers. The \textcolor{green}{$\Uparrow$} symbol represents a performance improvement over classical KD. The results are averaged over more than three runs.} \label{t3}
\centering
\resizebox{0.95\linewidth}{!}{
\begin{tabular}{|c|c|c|c|c|c|c|c|c|c|}
\bottomrule
\multicolumn{3}{|c|}{Distillation Stage} &
\multicolumn{2}{c|}{Response Stage} &
\multicolumn{2}{c|}{Feature Stage} &
\multicolumn{1}{c|}{Single Parallel Stage} &
\multicolumn{2}{c|}{Combination Stage} \\ 
\midrule
Accuracy & Teacher & Student & $\ast$KD$\bullet$\cite{IEEEexample:knowledgedistillation01} & $\ast$DKD$\bullet$\cite{IEEEexample:DKD} & $\ast$AT$\bullet$\cite{IEEEexample:knowledgedistillation04} & $\ast$CRD$\bullet$\cite{IEEEexample:CRD} & \textbf{SKD} & \textbf{RSKD} & {$\bullet$\textbf{RSKD}} \\ 
\midrule
top-1 &73.31 &69.75 &71.62 &71.86 &71.50 &71.83 &72.01\textcolor{green}{$\Uparrow$} &72.65\textcolor{green}{$\Uparrow$} &\textbf{72.93}\textcolor{green}{$\Uparrow$} \\
top-5 &91.42 &89.07 &90.33 &90.42 &90.40 &90.25 &90.56\textcolor{green}{$\Uparrow$} &91.20\textcolor{green}{$\Uparrow$} &\textbf{91.32}\textcolor{green}{$\Uparrow$} \\
\bottomrule
\end{tabular}}
\end{table*}

\begin{table*}
\caption{Top-1 and Top-5 accuracy (\%) of the student network MobileNet-V1 on ImageNet with teacher and student transfer across different architectures. ResNet-50 is used as the teacher network and MobileNet-V1 as the student network. The $\bullet$ symbol indicates the ensemble strategy using three branches, and the $\ast$ symbol indicates the cascading strategy using three layers. The \textcolor{green}{$\Uparrow$} symbol represents a performance improvement over classical KD. The results are averaged over more than three runs.} \label{t4}
\centering
\resizebox{0.95\linewidth}{!}{
\begin{tabular}{|c|c|c|c|c|c|c|c|c|c|}
\bottomrule
\multicolumn{3}{|c|}{Distillation Stage} &
\multicolumn{2}{c|}{Response Stage} &
\multicolumn{2}{c|}{Feature Stage} &
\multicolumn{1}{c|}{Single Parallel Stage} &
\multicolumn{2}{c|}{Combination Stage} \\ 
\midrule
Accuracy & Teacher & Student & $\ast$KD$\bullet$\cite{IEEEexample:knowledgedistillation01} & $\ast$DKD$\bullet$\cite{IEEEexample:DKD} & $\ast$AT$\bullet$\cite{IEEEexample:knowledgedistillation04} & $\ast$CRD$\bullet$\cite{IEEEexample:CRD} & \textbf{SKD} & \textbf{RSKD} & \textbf{$\bullet$RSKD$\dagger$} \\ 
\midrule
top-1 &76.16 &68.87 &68.98 &72.75 &70.16 &71.97 &72.91\textcolor{green}{$\Uparrow$} &73.86\textcolor{green}{$\Uparrow$} &\textbf{74.30}\textcolor{green}{$\Uparrow$} \\
top-5 &92.86 &88.76 &89.12 &91.65 &90.63 &91.20 &91.86\textcolor{green}{$\Uparrow$} &92.57\textcolor{green}{$\Uparrow$} &\textbf{92.94}\textcolor{green}{$\Uparrow$} \\
\bottomrule
\end{tabular}}
\end{table*}

\begin{table*}
\caption{Top-1 and Top-5 accuracy (\%) of the student network ResNet-18 on UCF101 with teacher and student transfer across the same architectures. ResNet-34 is used as the teacher network and ResNet-18 as the student network. The $\bullet$ symbol indicates the ensemble strategy using three branches, and the $\ast$ symbol indicates the cascading strategy using three layers. The \textcolor{green}{$\Uparrow$} symbol represents a performance improvement over classical KD. The results are averaged over more than three runs.} \label{t5}
\centering
\resizebox{0.95\linewidth}{!}{
\begin{tabular}{|c|c|c|c|c|c|c|c|c|c|}
\bottomrule
\multicolumn{3}{|c|}{Distillation Stage} &
\multicolumn{2}{c|}{Response Stage} &
\multicolumn{2}{c|}{Feature Stage} &
\multicolumn{1}{c|}{Single Parallel Stage} &
\multicolumn{2}{c|}{Combination Stage} \\ 
\midrule
Accuracy & Teacher & Student & $\ast$KD$\bullet$\cite{IEEEexample:knowledgedistillation01} & $\ast$DKD$\bullet$\cite{IEEEexample:DKD} & $\ast$AT$\bullet$\cite{IEEEexample:knowledgedistillation04} & $\ast$CRD$\bullet$\cite{IEEEexample:CRD} & \textbf{SKD} & \textbf{RSKD} & {$\bullet$\textbf{RSKD}} \\ 
\midrule
top-1 &38.67 &32.57 &34.12 &35.70 &34.65 &34.37 &35.86\textcolor{green}{$\Uparrow$} &36.48\textcolor{green}{$\Uparrow$} &\textbf{36.71}\textcolor{green}{$\Uparrow$} \\
top-5 &72.30 &61.48 &65.63 &67.19 &66.35 &66.72 &66.87\textcolor{green}{$\Uparrow$} &68.45\textcolor{green}{$\Uparrow$} &\textbf{68.91}\textcolor{green}{$\Uparrow$} \\
\bottomrule
\end{tabular}}
\end{table*}

\begin{table*}
\caption{Top-1 and Top-5 accuracy (\%) of the student network MobileNet-V1 on UCF101 with teacher and student transfer across different architectures. ResNet-50 is used as the teacher network and MobileNet-V1 as the student network. The $\bullet$ symbol indicates the ensemble strategy using three branches, and the $\ast$ symbol indicates the cascading strategy using three layers. The \textcolor{green}{$\Uparrow$} symbol represents a performance improvement over classical KD. The results are averaged over more than three runs.} \label{t6}
\centering
\resizebox{0.95\linewidth}{!}{
\begin{tabular}{|c|c|c|c|c|c|c|c|c|c|}
\bottomrule
\multicolumn{3}{|c|}{Distillation Stage} &
\multicolumn{2}{c|}{Response Stage} &
\multicolumn{2}{c|}{Feature Stage} &
\multicolumn{1}{c|}{Single Parallel Stage} &
\multicolumn{2}{c|}{Combination Stage} \\ 
\midrule
Accuracy & Teacher & Student & $\ast$KD$\bullet$\cite{IEEEexample:knowledgedistillation01} & $\ast$DKD$\bullet$\cite{IEEEexample:DKD} & $\ast$AT$\bullet$\cite{IEEEexample:knowledgedistillation04} & $\ast$CRD$\bullet$\cite{IEEEexample:CRD} & \textbf{SKD} & \textbf{RSKD} & \textbf{$\bullet$RSKD$\dagger$} \\ 
\midrule
top-1 &40.62 &36.71 &37.54 &38.26 &37.60 &37.92 &38.52\textcolor{green}{$\Uparrow$} &39.17\textcolor{green}{$\Uparrow$} &\textbf{40.06}\textcolor{green}{$\Uparrow$} \\
top-5 &76.61 &65.18 &67.94 &70.25 &68.46 &69.11 &70.39\textcolor{green}{$\Uparrow$} &72.06\textcolor{green}{$\Uparrow$} &\textbf{73.69}\textcolor{green}{$\Uparrow$} \\
\bottomrule
\end{tabular}}
\end{table*}

\subsection{Comparison of Training Efficiency}

\begin{figure}
\centering
\includegraphics[width=0.5\textwidth]{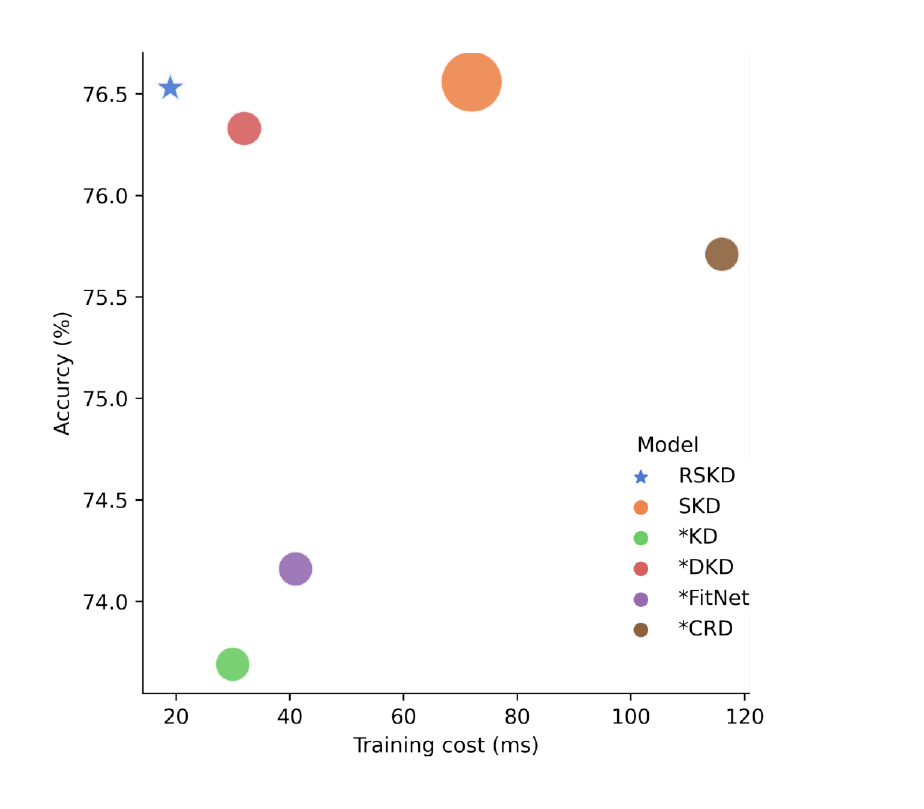}
\label{fig:fifth}
\caption{Comparison of accuracy and training time per batch on the CIFAR-100 dataset. The student network is ResNet8$\times$4 and the teacher network is ResNet32$\times$4.}
\label{fig:fifth}
\vspace{-0.6cm}
\end{figure}

\begin{table}
\caption{Results of student networks on CIFAR-100, where teachers and students transfer across different architectures. The table includes results obtained using a cascading strategy with three layers, denoted by $\ast$.} \label{t10}
\centering
\resizebox{0.6\linewidth}{!}{
\begin{tabular}{cccc}
\midrule
\multirow{1}{*}{}
& Top-1 & Time & Params\\
\midrule
\multirow{1}{*}{$\ast$KD\cite{IEEEexample:knowledgedistillation01}}
& 73.69 & 30 & 0\\
\multirow{1}{*}{$\ast$DKD\cite{IEEEexample:DKD}}
& 76.33 & 32 & 0\\
\multirow{1}{*}{$\ast$FITNET\cite{IEEEexample:knowledgedistillation02}}
& 74.16 & 41 & 46.5K\\
\multirow{1}{*}{$\ast$CRD\cite{IEEEexample:CRD}}
& 75.71 & 116 & 31.7M\\
\midrule
\multirow{1}{*}{\textbf{SKD}}
& \textbf{76.56} & 72 & 62.3K\\
\multirow{1}{*}{\textbf{RSKD}}
& 76.53 & \textbf{19} & 42.3K\\
\midrule
\end{tabular}}
\vspace{-0.6cm}
\end{table}

We evaluate the training costs of state-of-the-art distillation methods and demonstrate that RSKD is highly efficient. As illustrated in \cref{fig:fifth} and \cref{t10}, RSKD achieves an optimal balance between model performance and training costs. The training accuracy of RSKD surpasses the most advanced baseline distillation models while maintaining minimal training time and the same model size. Compared to SKD, RSKD is more efficient at only one-third of the size of SKD and saves nearly three times the training time.

\begin{figure*}
  \begin{subfigure}{0.24\linewidth}
    \centering
    \includegraphics[width=0.8\textwidth]{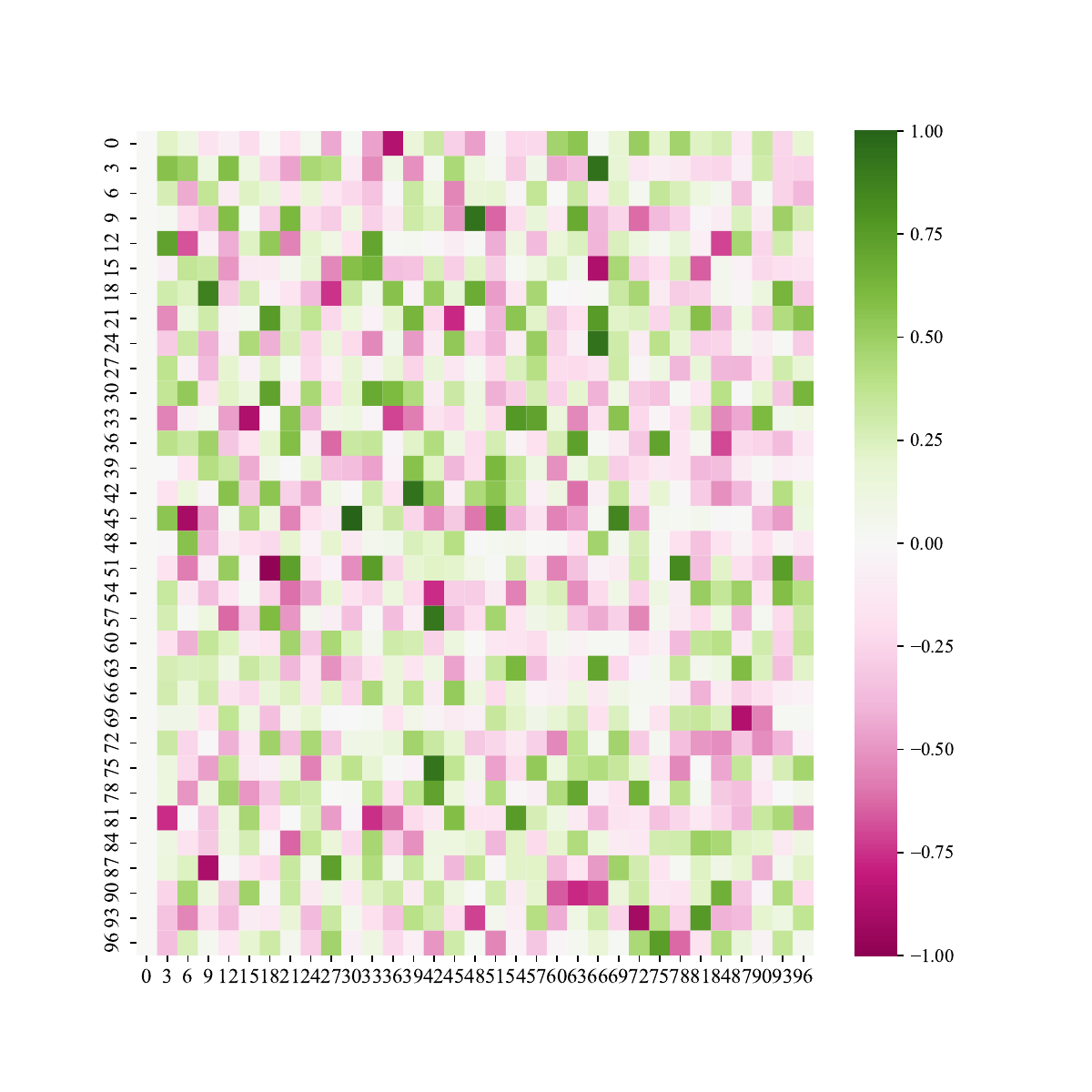}
	\caption{SKD$\circ$.}
    \label{fig:sixth-a}
  \end{subfigure}
  \begin{subfigure}{0.24\linewidth}
    \centering
    \includegraphics[width=0.8\textwidth]{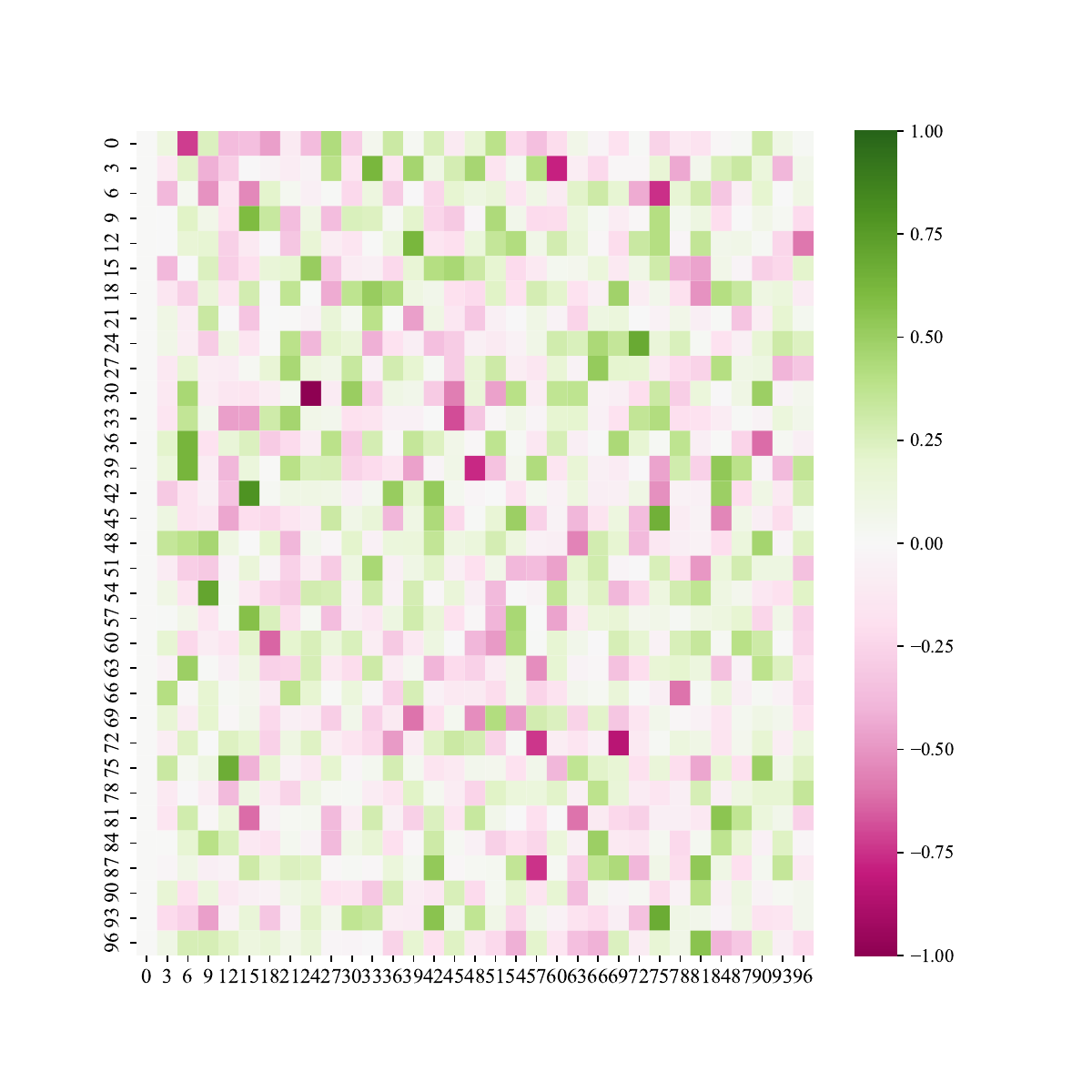}
    \caption{SKD$\star$.}
    \label{fig:sixth-b}
  \end{subfigure}
  \begin{subfigure}{0.24\linewidth}
    \centering
    \includegraphics[width=0.8\textwidth]{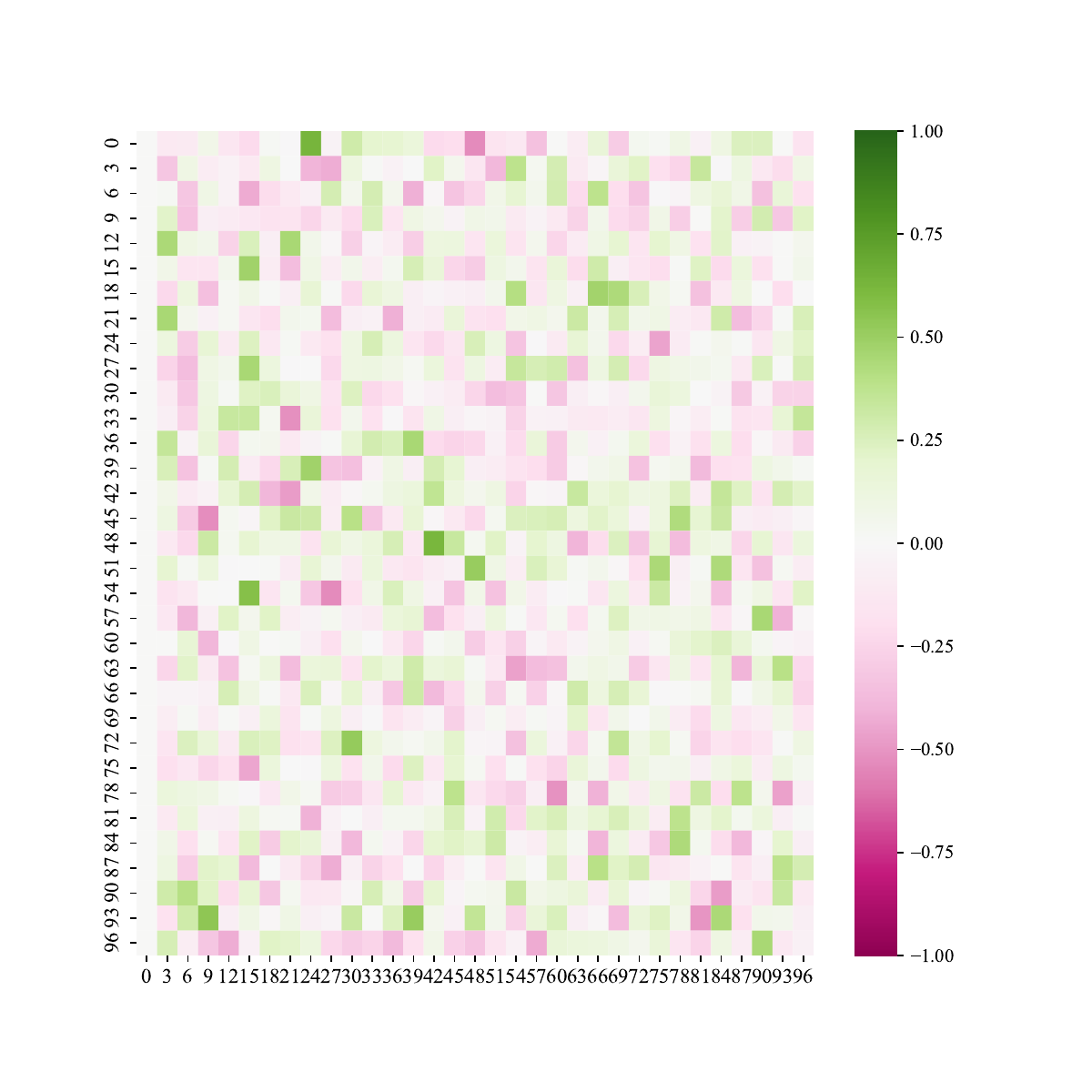}
	\caption{SKD$\dagger$.}
    \label{fig:sixth-c}
  \end{subfigure}
  \begin{subfigure}{0.24\linewidth}
    \centering
    \includegraphics[width=0.8\textwidth]{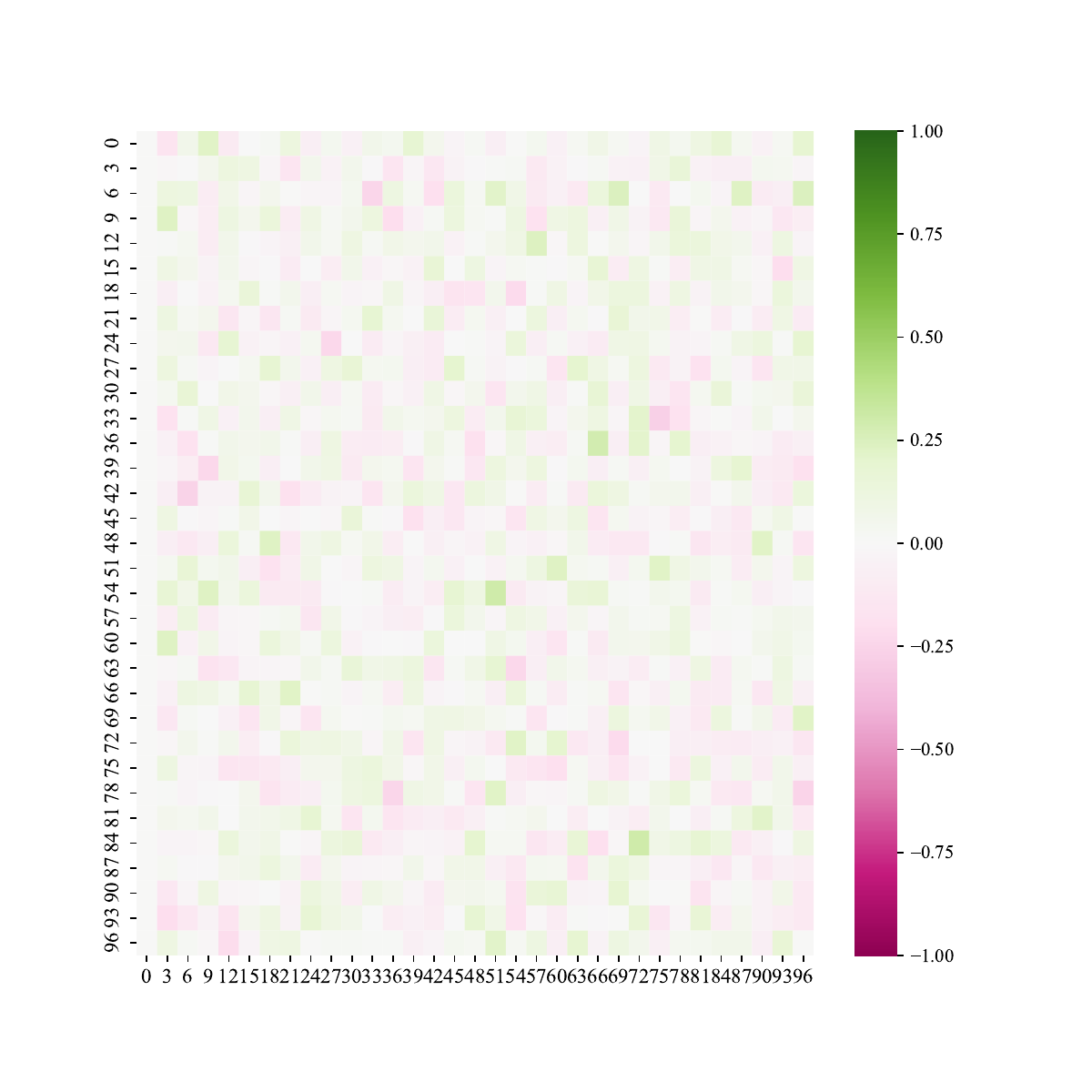}
    \caption{RSKD.}
    \label{fig:sixth-d}
  \end{subfigure}
\caption{Comparison of correlation matrices between teacher and student logits. We use ResNet56 as the teacher and ResNet20 as the student on CIFAR-100. The correlation matrices for different stages are shown: (a) Response-based stage ($\circ$), (b) Feature-based stage ($\star$), (c) Relation-based stage ($\dagger$), and (d) RSKD.}
\vspace{-0.6cm}
\label{fig:sixth}
\end{figure*}

\begin{figure*}
\centering
	\begin{subfigure}{0.24\linewidth}
		\centering
		\includegraphics[width=0.9\linewidth]{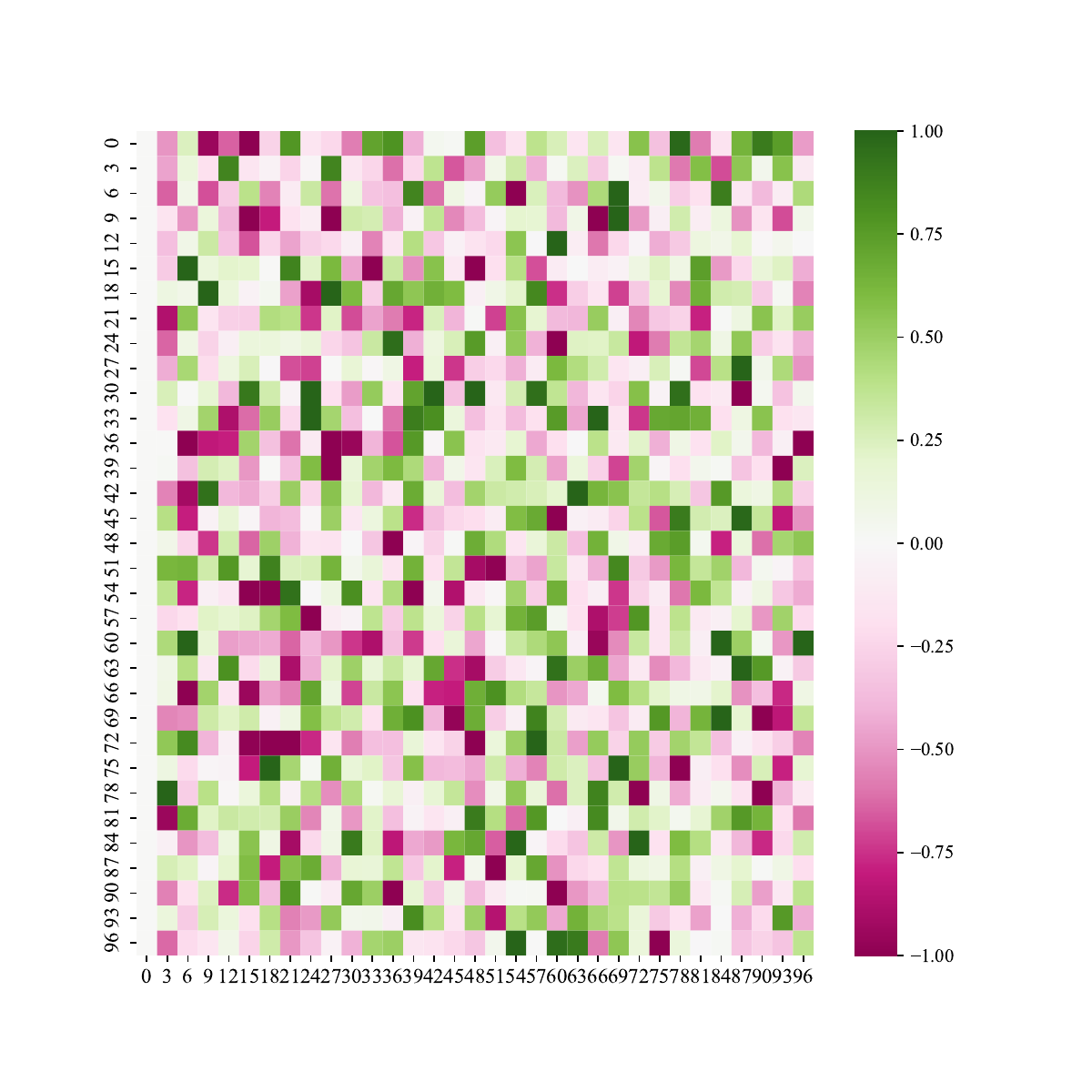}
		\caption{SKD$\circ$}
	\end{subfigure}
	\centering
	\begin{subfigure}{0.24\linewidth}
		\centering
		\includegraphics[width=0.9\linewidth]{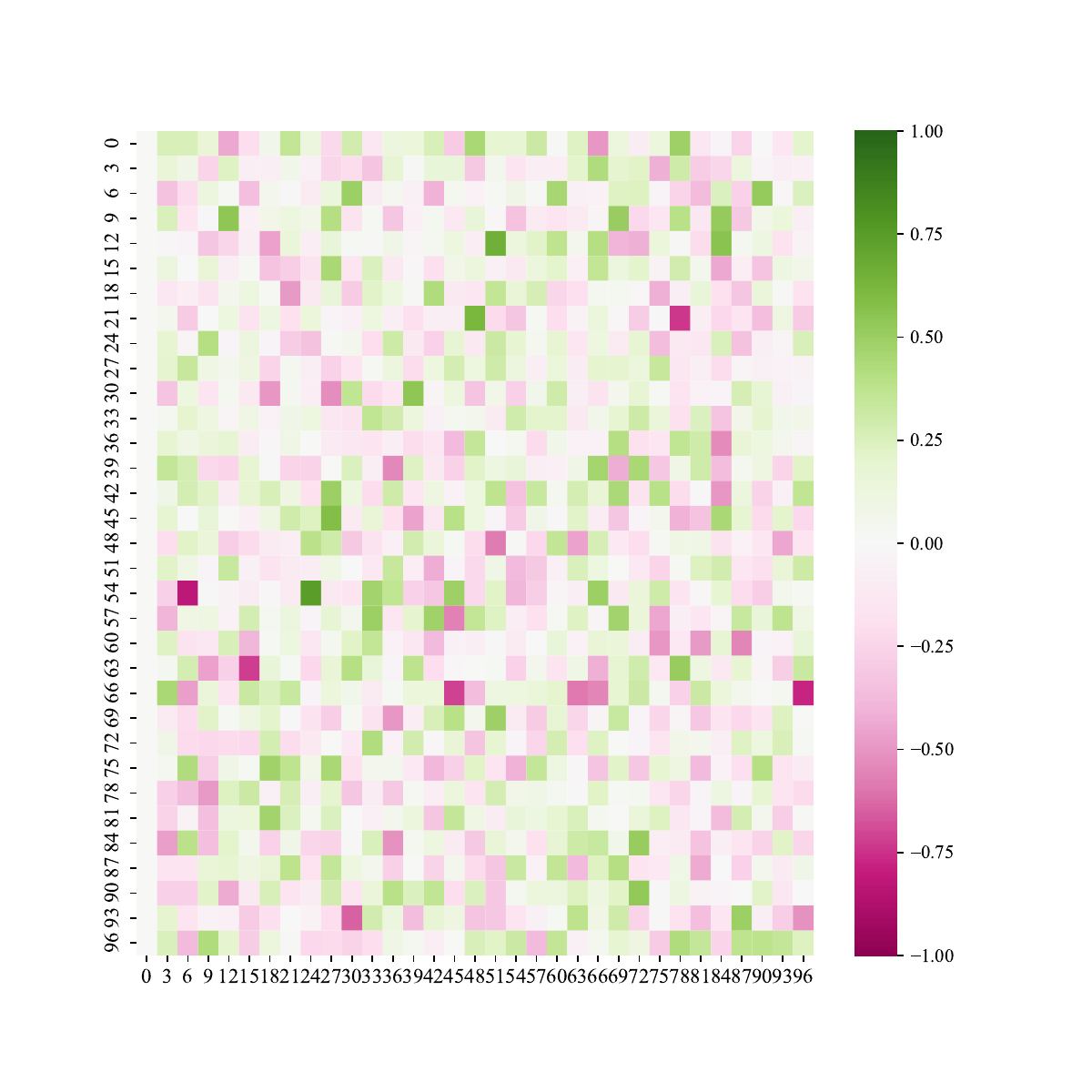}
		\caption{SKD$\star$}
	\end{subfigure}
	\centering
	\begin{subfigure}{0.24\linewidth}
		\centering
		\includegraphics[width=0.9\linewidth]{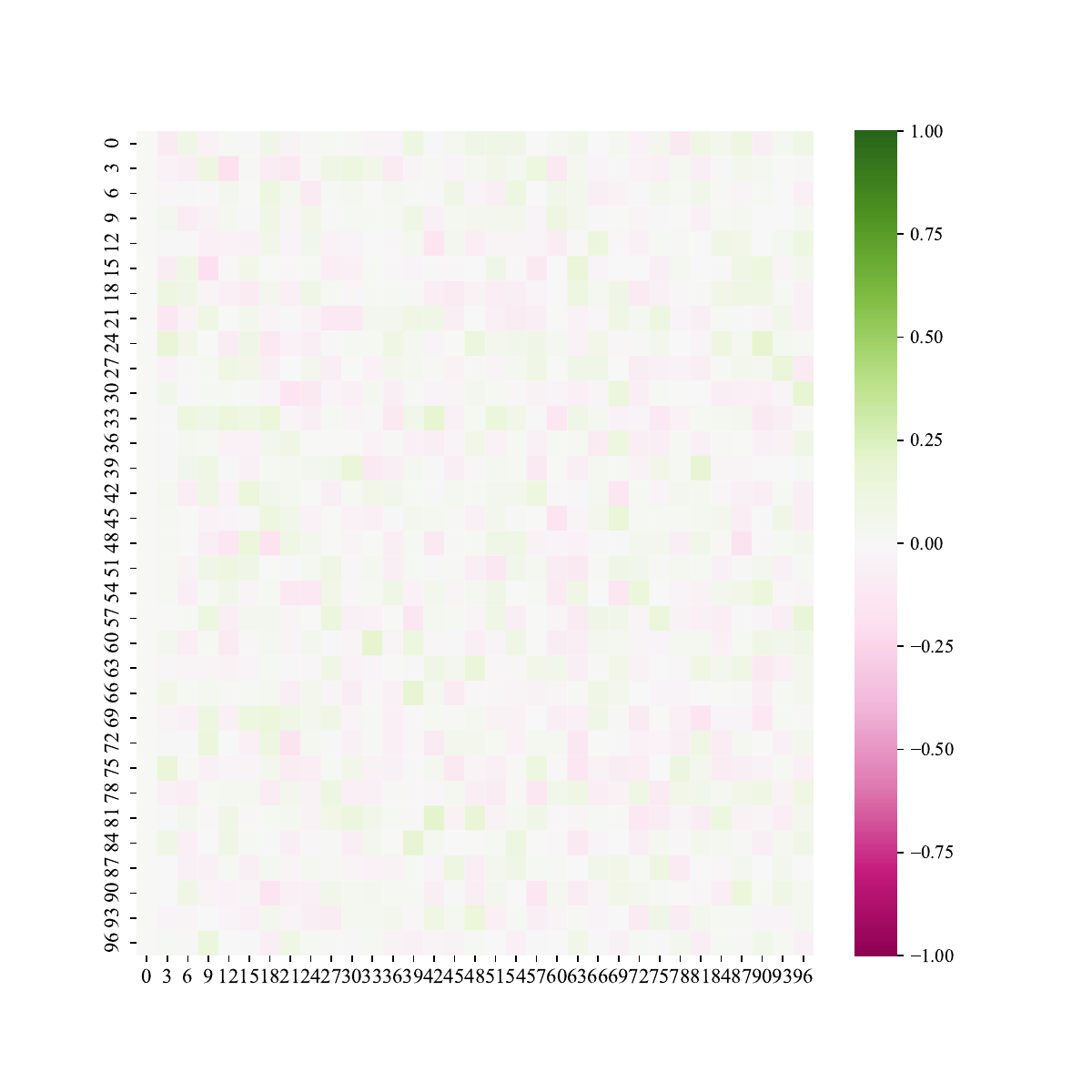}
		\caption{RSKD$\dagger$}
	\end{subfigure}
	\caption{Differences in correlation matrices between teacher and student logits with the same network structure. (a) SKD$\circ$ indicates that the response-based stage is used only; (b) SKD$\star$ indicates that the feature-based stage is used only; (c) RSKD$\dagger$ with CS structure has the smallest difference, indicating a significant match between teachers and students compared to the above two methods. We set ResNet50 as the teacher and MobileNetV1 as the student on ImageNet. $\dagger$ indicates that the relation-based stage is removed.}
\vspace{-0.6cm}
\label{f52}
\end{figure*}

\subsection{Visualization of Correlation}

We visualize the differences in correlation matrices between teacher and student logits for various distillation methods to demonstrate the effectiveness of RSKD in preserving the teacher's knowledge. In CIFAR-100, we compare four students: SKD$\circ$, SKD$\star$, SKD$\dagger$, and RSKD, as shown in \cref{fig:sixth}. The results indicate that RSKD achieves the most consistent correlation between teachers and students using the CS structure, resulting in highly similar logits with minimal discrepancies. This suggests that RSKD is a practical approach for preserving the teacher's knowledge in the student.
% In CIFAR-100,we compute and visualize the differences in correlation matrices between teacher and student logits for four students: SKD$\circ$, SKD$\star$, SKD$\dagger$, and RSKD in \cref{fig:sixth}. RSKD shows the most consistent correlation between teachers and students using CS, enabling students to generate highly similar logits, as evidenced by the minimal discrepancies between teacher and student logits. This suggests that RSKD is a practical approach for producing logits that closely resemble the teacher's, providing reliable results.

In ImageNet, we further evaluate the correlation between teacher and student logits for different combinations of teacher-student networks. \cref{f52} demonstrate that RSKD minimizes the differences in correlation matrices for different teacher-student networks, indicating that RSKD can preserve the teacher's knowledge across different network architectures. Therefore, our method can generate student logits that closely resemble the teacher's, providing reliable results.
% In ImageNet, we calculate and visualize the differences in correlation matrices between the logIts of different combinations of teacher and student networks. As shown in \cref{f51} and \cref{f52}, the RSKD method best minimizes correlation differences between the same or different teacher-student networks.

These results show that preserving the correlation between teacher and student logits is crucial for effective knowledge transfer in distillation. The preservation of correlation ensures that the student network reproduces the same predictions as the teacher network, even if the student network is smaller or has a different architecture than the teacher network. The correlation visualization provides further evidence that RSKD can effectively preserve the teacher's knowledge in the student network.

\begin{table}
\caption{Ablation study of RSKD on CIFAR-100, showing the impact of different components. The first column indicates whether cascading is used ({\textcircled{\tiny{CA}}}), while the second column indicates whether the Distilled Features (DF) method is employed ({\textcircled{\tiny{DF}}}). The following columns represent the effect of using different backbone networks for the correlation shaping (CS) module: {\textcircled{\tiny{C-r}}} for the response-based backbone network, {\textcircled{\tiny{C-f}}} for the feature-based backbone network, and {\textcircled{\tiny{C-e}}} for the relation-based backbone network. The results are reported in terms of the top-1 accuracy (\%) of the student network.} \label{t5}
\centering
\resizebox{.9\linewidth}{!}{
\begin{tabular}{|cc|ccccc|c|}
\bottomrule
\multicolumn{2}{|c|}{{\multirow{2}{*}{Teacher Student}}} &
\multicolumn{5}{c|}{Ablation} &
\multicolumn{1}{c|}{{\multirow{2}{*}{Top-1}}} \\ 
\multicolumn{2}{|c|}{~}&\Large{\textcircled{\scriptsize{CA}}}\Large & \Large{\textcircled{\scriptsize{DF}}}\Large  & \Large{\textcircled{\scriptsize{C-r}}}\Large & \Large{\textcircled{\scriptsize{C-f}}}\Large & \Large{\textcircled{\scriptsize{C-e}}}\Large & \\ 
\midrule
\multicolumn{2}{|c|}{\multirow{4}{*}{ResNet56 ResNet20}}& - & - & - & - & - &69.82 \\
	%\cline{3-6}
	\multicolumn{2}{|c|}{~}&\checkmark & - & - & - & - &71.76 \\
	%\cline{3-6}
	\multicolumn{2}{|c|}{~}&\checkmark &\checkmark & - & - & - &\textbf{71.82} \\
	%\cline{3-6}
	\multicolumn{2}{|c|}{~}&\checkmark &\checkmark &\checkmark &- & - &71.82 \\
	%\cline{3-6}
	\multicolumn{2}{|c|}{~}&\checkmark &\checkmark &- &\checkmark & - &71.79 \\
	%\cline{3-6}
	\multicolumn{2}{|c|}{~}&\checkmark &\checkmark &- &- &\checkmark &71.63 \\
\midrule
\multicolumn{2}{|c|}{\multirow{4}{*}{ResNet110 ResNet32}}& - & - & - & - & - &71.56 \\
	%\cline{3-6}
	\multicolumn{2}{|c|}{~}&\checkmark & - & - & - & - &73.92 \\
	%\cline{3-6}
	\multicolumn{2}{|c|}{~}&\checkmark &\checkmark & - & - & - &74.06 \\
%\cline{3-6}
	\multicolumn{2}{|c|}{~}&\checkmark &\checkmark &\checkmark &- & - &\textbf{74.21} \\
%\cline{3-6}
	\multicolumn{2}{|c|}{~}&\checkmark &\checkmark &- &\checkmark & - &74.11\\
	%\cline{3-6}
	\multicolumn{2}{|c|}{~}&\checkmark &\checkmark &- &- &\checkmark &74.07 \\
\midrule
\multicolumn{2}{|c|}{\multirow{4}{*}{ResNet32×4 ResNet8×4}}& - & - & - & - & - &73.67 \\
	%\cline{3-6}
	\multicolumn{2}{|c|}{~}&\checkmark & - & - & - & - &75.89 \\
	%\cline{3-6}
	\multicolumn{2}{|c|}{~}&\checkmark &\checkmark & - & - & - &76.20 \\
	%\cline{3-6}
	\multicolumn{2}{|c|}{~}&\checkmark &\checkmark &\checkmark &- & - &\textbf{76.53} \\
	%\cline{3-6}
	\multicolumn{2}{|c|}{~}&\checkmark &\checkmark &- &\checkmark & - &76.32 \\
	%\cline{3-6}
	\multicolumn{2}{|c|}{~}&\checkmark &\checkmark &- &- &\checkmark &76.21 \\
\midrule
\multicolumn{2}{|c|}{\multirow{4}{*}{ResNet50 MobileNetV2}}& - & - & - & - & - &68.41 \\
	%\cline{3-6}
	\multicolumn{2}{|c|}{~}&\checkmark & - & - & - & - &70.32 \\
	%\cline{3-6}
	\multicolumn{2}{|c|}{~}&\checkmark &\checkmark & - & - & - &70.39\\
	%\cline{3-6}
	\multicolumn{2}{|c|}{~}&\checkmark &\checkmark&\checkmark &- & - &\textbf{70.55}\\
	%\cline{3-6}
	\multicolumn{2}{|c|}{~}&\checkmark &\checkmark &- &\checkmark &- &70.43  \\
\bottomrule
\end{tabular}}
\vspace{-0.59cm}
\end{table}

\begin{table}
\caption{Ablation study of RSKD on UCF101, showing the impact of different components. The first column indicates whether cascading is used ({\textcircled{\tiny{CA}}}), while the second column indicates whether the Distilled Features (DF) method is employed ({\textcircled{\tiny{DF}}}). The following columns represent the effect of using different backbone networks for the correlation shaping (CS) module: {\textcircled{\tiny{C-r}}} for the response-based backbone network, {\textcircled{\tiny{C-f}}} for the feature-based backbone network, and {\textcircled{\tiny{C-e}}} for the relation-based backbone network. The results are reported in terms of the top-1 accuracy (\%) of the student network.} \label{t7}
\centering
\resizebox{.8\linewidth}{!}{
\begin{tabular}{|cc|ccccc|c|}
\bottomrule
\multicolumn{2}{|c|}{{\multirow{2}{*}{Teacher Student}}} &
\multicolumn{5}{c|}{Ablation} &
\multicolumn{1}{c|}{{\multirow{2}{*}{Top-1}}} \\ 
\multicolumn{2}{|c|}{~}&\Large{\textcircled{\scriptsize{CA}}}\Large & \Large{\textcircled{\scriptsize{DF}}}\Large  & \Large{\textcircled{\scriptsize{C-r}}}\Large & \Large{\textcircled{\scriptsize{C-f}}}\Large & \Large{\textcircled{\scriptsize{C-e}}}\Large & \\ 
\midrule
\multicolumn{2}{|c|}{\multirow{4}{*}{ResNet34 ResNet18}}& - & - & - & - & - &32.57 \\
	%\cline{3-6}
	\multicolumn{2}{|c|}{~}&\checkmark & - & - & - & - &33.69 \\
	%\cline{3-6}
	\multicolumn{2}{|c|}{~}&\checkmark &\checkmark & - & - & - &34.12 \\
	%\cline{3-6}
	\multicolumn{2}{|c|}{~}&\checkmark &\checkmark &\checkmark &- & - &\textbf{36.48} \\
	%\cline{3-6}
	\multicolumn{2}{|c|}{~}&\checkmark &\checkmark &- &\checkmark & - &36.10 \\
	%\cline{3-6}
	\multicolumn{2}{|c|}{~}&\checkmark &\checkmark &- &- &\checkmark &35.64 \\
\midrule
\multicolumn{2}{|c|}{\multirow{4}{*}{ResNet50 MobileNetV1}}& - & - & - & - & - &36.71 \\
	%\cline{3-6}
	\multicolumn{2}{|c|}{~}&\checkmark & - & - & - & - &37.24 \\
	%\cline{3-6}
	\multicolumn{2}{|c|}{~}&\checkmark &\checkmark & - & - & - &37.86 \\
	%\cline{3-6}
	\multicolumn{2}{|c|}{~}&\checkmark &\checkmark &\checkmark &- & - &\textbf{39.17} \\
	%\cline{3-6}
	\multicolumn{2}{|c|}{~}&\checkmark &\checkmark &- &\checkmark & - &38.52 \\
\bottomrule
\end{tabular}}

\end{table}

\begin{table}

\caption{Ablation study of DF in RSKD on ImageNet, showing the impact of different K values. The results are reported in terms of the top-1 accuracy (\%) of the student network.} \label{t8}
\centering
\resizebox{.8\linewidth}{!}{
\begin{tabular}{|cc|ccccccc|c|}
\bottomrule
\multicolumn{2}{|c|}{{\multirow{2}{*}{Teacher Student}}} &
\multicolumn{7}{c|}{Ablation} &
\multicolumn{1}{c|}{{\multirow{2}{*}{Top-1}}} \\ 
\multicolumn{2}{|c|}{~}&\Large{\textcircled{\scriptsize{CA}}}\Large&\Large{\textcircled{\scriptsize{C-r}}}\Large&2&4&8&16&32& \\ 
\midrule
\multicolumn{2}{|c|}{\multirow{4}{*}{ResNet34 ResNet18}}&\checkmark&\checkmark&\checkmark & - & - & - & - &69.83 \\
	%\cline{3-6}
	\multicolumn{2}{|c|}{~}&\checkmark&\checkmark&-&\checkmark& - & - & - &71.41 \\
	%\cline{3-6}
	\multicolumn{2}{|c|}{~}&\checkmark&\checkmark&-&-& \checkmark  & - & - &72.12 \\
	%\cline{3-6}
	\multicolumn{2}{|c|}{~}&\checkmark&\checkmark&-&- &-&\small{\textcircled{\scriptsize{\checkmark}}}\small & - &\textbf{72.65}  \\
	%\cline{3-6}
	\multicolumn{2}{|c|}{~}&\checkmark&\checkmark&-&-&- &- &\checkmark &72.48 \\
\midrule
\multicolumn{2}{|c|}{\multirow{4}{*}{ResNet50 MobileNetV1}}&\checkmark&\checkmark&\checkmark& - & - & - & - &72.68 \\
	%\cline{3-6}
	\multicolumn{2}{|c|}{~}&\checkmark&\checkmark&- &\checkmark& - & - & - &73.43 \\
	%\cline{3-6}
	\multicolumn{2}{|c|}{~}&\checkmark&\checkmark&-&- &\small{\textcircled{\scriptsize{\checkmark}}}\small& - & - &\textbf{73.86} \\
	%\cline{3-6}
	\multicolumn{2}{|c|}{~}&\checkmark&\checkmark&-&-&-&\checkmark& - &73.57 \\
	%\cline{3-6}
	\multicolumn{2}{|c|}{~}&\checkmark&\checkmark&-&-&- &-&\checkmark&73.21 \\
\bottomrule
\end{tabular}}
\vspace{-0.3cm}
\end{table}

\subsection{Ablation Study}

We conducted ablation experiments on the RSKD method to evaluate the impact of cascade operation, DF method, CS structure, and K value on the accuracy of CIFAR-100, ImageNet and UCF101 datasets. The results are presented in \cref{t5}, \cref{t7}, and \cref{t8}.
% In \cref{t5} and \cref{t6}, we conduct ablation experiments on our proposed RSKD method to assess the impact of cascade operation, DF method, and CS structure on the accuracy of CIFAR-100 and ImageNet datasets.

Our experiments demonstrate that the response-based structure of the backbone network achieves optimal results in general. For small datasets with the same teacher-student model structure, the cascade operation and DF method have a more significant effect on the model performance. In contrast, reasonable CS structures have the same or slightly improved performance. However, the CS structure is more effective for models with larger datasets or different teacher-student model structures. This is because models with CS structures are less prone to overfitting, especially with a large amount of data or heterogeneous teacher-student networks.

The value of K is simultaneously determined using the grid search method. For teacher-student networks with the same structure, in the case of an optimal solution, the value of K tends to be larger. Conversely, for teacher-student networks with different structures, in the case of an optimal solution, the value of K is relatively smaller. This disparity arises from the fact that models with a homogeneous structure are more influenced by the first K channels in each layer, whereas heterogeneous models are less affected by the range of K values.
% The response-based structure of the backbone network achieves optimum results in general. For small datasets with the same teacher-student model structure, the cascade operation and DF method have a more significant effect on the performance of the models. In contrast, reasonable CS structures have the same or slightly improved performance. However, the CS structure is more effective for models with larger datasets or different teacher-student model structures. This is because models with CS structures are less prone to overfitting given a large amount of data or heterogeneous teacher-student networks.

In summary, the RSKD method with the appropriate combination of cascade operation, DF method, and CS structure can significantly improve the model's accuracy, particularly for large or diverse datasets.

%-------------------------------------------------------------------------
\section{Conclusion}

In this paper, we have proposed novel weakly supervised teacher-student architectures, SKD and RSKD, that transform the KD process into a substage learning process, improving the quality of pseudo labels. We extensively investigate the relationship between the type of sub-stage learning process and the teacher-student structure, and demonstrate the validity of our method on the video classification task of CIFAR-100, ImageNet package dataset and UCF101 real dataset. Our analysis of the design methodology of the multi-branch substage composite structure and the optimization of the corresponding loss function has further validated the human-inspired design of the teacher-student network structure and substage learning process, leading to improved performance.

Importantly, our work is highly relevant to label-efficient learning on video data, which aims to explore new methods for video labeling and analysis. Our proposed SKD and RSKD methods offer promising approaches to improve the quality of pseudo labels in video classification tasks, which is a challenging problem due to the complexity and high dimensionality of video data. Inspired by human staged learning strategies, our approach offers an intuitive framework that is not limited to specific computer vision tasks.   We believe that our strategies can be further applied to other computer video tasks such as video segmentation, single object tracking, and multiple object tracking.    These tasks have gained considerable attention in the computer vision community, and we believe that our approach can be leveraged to enhance their performance as well.   Overall, our work contributes to advancing the field of weakly supervised learning and can have significant practical implications in real-world applications.
% We believe our strategies will enhance the efficacy of KD techniques when applied to large datasets and intricate models. In the future, we plan to utilize our SKD and RSKD methods for various computer vision tasks, such as object detection and semantic segmentation, to investigate their potential further.

\bibliographystyle{IEEEtran}
\bibliography{TCSVT}
\begin{IEEEbiography}[{\includegraphics[width=1in,height=1.25in,clip,keepaspectratio]{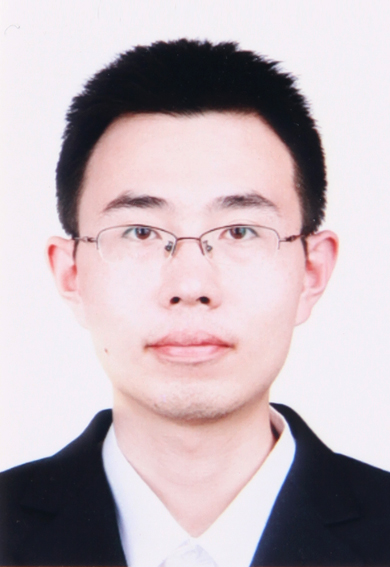}}]{Chao Wang} 					                                          
received a B.E. degree in Process Equipment and Control Engineering from Jiangnan University in 2007, and an M.S. degree in Computer Application Technology from Northeast University in 2010. In the same year, he joined the China Academy of Railway Sciences and is currently an Assistant Researcher. Since assuming this position in 2012, he has been in charge of or involved in obtaining 8 invention patents and 3 utility model patents. His published papers cover interdisciplinary research topics in the fields of big data, neural networks, machine learning, and rail transportation. His research interests include intelligent rail transit systems, big data systems, cloud computing, computer vision, and graph neural networks related to machine learning.

Chao Wang has led or participated in several key projects in China's rail transportation industry, achieving significant socioeconomic benefits. In 2021, his team won the second prize of the Science and Technology Progress of Beijing Rail Transit Society Award. The project he participated in received the first prize of the China Academy of Railway Sciences Award in 2020. In 2018, the project he led received the Innovation Award of the Communication Signal Research Institute. An invention patent he participated in won the China Excellent Patent Award in 2017.
\end{IEEEbiography}

\begin{IEEEbiography}[{\includegraphics[width=1in,height=1.25in,clip,keepaspectratio]{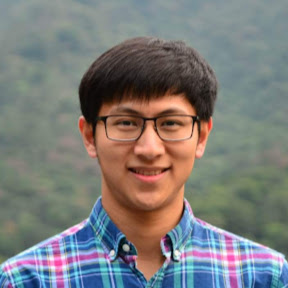}}]{Zheng Tang }
(M’16) earned his B.Sc. (Eng.) degree with Honors from the joint program by Beijing University of Posts and Telecommunications and Queen Mary University of London in 2014. He further obtained his M.S. and Ph.D. degrees in Electrical and Computer Engineering from the University of Washington in 2016 and 2019, respectively.

In 2019, Dr. Tang joined Amazon as an Applied Scientist for the Amazon One team, serving there until 2021. Currently, he is a Senior Data Science Engineer at NVIDIA's Metropolis division, a position he has held since 2021. He has already had 3 U.S. patents and 17 peer-reviewed publications. His research interests include intelligent transportation systems, object tracking, re-identification, and other topics in the computer vision and machine learning realm.

Dr. Tang has been an Associate Editor for the IEEE Transactions on Circuits and Systems for Video Technology since 2021, and an Organizing Committee Member for the AI City Challenge Workshops in conjunction with CVPR since 2020. He will also serve as the Challenge Chair for AVSS 2023, and previously served as an Area Chair for MLSP 2021. He received the Best AE Award of T-CSVT in 2021. A team he led triumphed in the 2nd AI City Challenge Workshop in CVPR 2018, securing the first rank in two tracks. Additionally, his paper was a finalist for two Best Student Paper Awards at ICPR 2016.
\end{IEEEbiography}

\end{document}